\documentclass[sigconf]{acmart}
\setcopyright{rightsretained}
\copyrightyear{2026}
\acmYear{2026}
\acmConference[Preprint]{Preprint}{2026}{under review}
\acmBooktitle{}
\acmDOI{}
\acmISBN{}
\newcounter{preprintauthorseparator}
\newcommand{\preprintauthorseparator}{%
  \stepcounter{preprintauthorseparator}%
  \ifnum\value{preprintauthorseparator}=4,\\%
  \else\ifnum\value{preprintauthorseparator}=8,\\%
  \else,\space\fi\fi}
\makeatletter
\@ifundefined{hyxmp@parse@acmart}
  {\PackageError{paper-arxiv}{Cannot disable ACM publication metadata}{Check the hyperxmp version.}}
  {\let\hyxmp@parse@acmart\relax}
\@printpermissionfalse
\patchcmd{\@mkauthors@iii}{\par}{\relax}
  {}{\PackageError{paper-arxiv}{Cannot format the first author}{Check the acmart author layout.}}
\patchcmd{\@mkauthors@iii}{\par}{\preprintauthorseparator}
  {}{\PackageError{paper-arxiv}{Cannot format the author list}{Check the acmart author layout.}}
\makeatother

\usepackage{multirow}
\usepackage{tabularx}
\usepackage{adjustbox}
\usepackage{booktabs}
\usepackage{tcolorbox}
\usepackage{siunitx}
\usepackage{subcaption}
\usepackage{tikz}
\usepackage{xcolor}
\usetikzlibrary{shapes.geometric, arrows.meta, positioning, shadows, fit, backgrounds, calc} 
\usepackage{pgfplots}
\usepgfplotslibrary{fillbetween}
\pgfplotsset{compat=1.18} %
\pgfplotsset{filter discard warning=false}
\pgfplotsset{width=\columnwidth} %
\usepackage{threeparttable}
\definecolor{metablue}{RGB}{0,100,224}
\usepackage{nicefrac}

\newcommand{\NE}{\mathrm{NE}}
\newcommand{\COST}{\mathrm{COST}}
\newcommand{\Pareto}{\mathcal{P}}

\begin{document}

\title{Data-Efficient Scaling-Law Discovery for Recommendation Systems via Coupled Hyperparameter Scaling} 
\author{John~Martabano~Landy}
\authornote{Joint first authors.}
\author{Louis~C.~Tiao}
\authornotemark[1]
\author{Yao~Xuan}
\authornotemark[1]
\author{Luc~Kien~Hang}
\author{Rupinder~P.~Khandpur}
\author{Arvind~M.~Vepa}
\author{Srinidhi~Raghavendran}
\author{Anirudh~Agrawal}
\author{Xiaoyi~Leo~Liu}
\author{Jiang~Liu}
\author{Peggy~Yao}
\author{Alireza~Vahdatpour}
\author{Djordje~Gligorijevic}
\affiliation{\institution{Meta}\city{}\country{}}
\renewcommand{\shortauthors}{Landy, Tiao, Xuan, et al.}

\begin{abstract} 
    Large-scale ranking and recommendation models of Web platforms benefit from increased compute much as large language models do, and their scaling behavior is broadly power-law in form. Yet ranking models have heterogeneous architectures which makes their scaling law less straightforward than that of LLMs. We address how to discover the scaling law of an arbitrary ranking model efficiently. We introduce a scaling layer, which groups primitive architecture parameters into semantically coherent groups and defines how they co-vary, inducing a mapping from a low-dimensional space of hyper-hyperparameters into the full architecture space. On top of it, Axia fits scaling-informed surrogates, selects configurations by a posterior-based acquisition criterion, and returns a Pareto frontier from which per-model operating points are validated -- replacing exhaustive sweeps with a small number of informative runs. A fleet-level evaluation protocol, computed on a fixed representative model set under matched trial budgets, makes refinements to either component measurable. Our contribution is an automated system that learns model-specific complexity--quality relationships and gives uncertainty-aware guidance for allocating compute across a heterogeneous model ecosystem. Across 34 model iterations in two development cycles on Meta's ads ranking and retrieval system, the workflow yielded a mean 0.66\% offline normalized-entropy improvement and 0.36\% cumulative online lift.

\end{abstract} 


\maketitle 

\section{Introduction}
Web platforms deliver content and advertising through ranking and retrieval models. These models decide what each request returns, under latency budgets set by the surfaces they serve, over user interaction data generated continuously. A platform of this kind operates a large number of models, each bound to a different surface, objective, feature set, and serving budget. The aggregate quality the platform can deliver is therefore set less by any individual model than by how finite serving compute is distributed across the fleet - and that allocation is currently made by hand, one model at a time, at a cost of days to weeks of training per decision. The challenge we address is how platforms should sustainably maximize model performance under an allocated compute budget.

Scaling laws have become an important tool for understanding how machine-learning performance changes with model capacity, data, and computation. In large language models, these relationships have proved sufficiently regular to guide training decisions and compute allocation~\citep{kaplan2020scaling}. Motivated by this success, recent work has begun studying analogous relationships in recommendation and ranking systems~\citep{ardalani2022understanding,zhang2024wukong,shen2024optimizing}. Our results reinforce this emerging picture: despite substantial differences in model structure and input representation, ranking models exhibit similarly predictable improvements as they scale.

Turning this predictability into a practical scaling strategy remains a challenge. Ranking models combine heterogeneous features, learned embeddings, feature-interaction modules, and sequences of user events, so their capacity is distributed across several architectural dimensions rather than summarized by parameter count alone. Those dimensions are coupled: increasing one component without adjusting the others can introduce a new bottleneck, use compute inefficiently, or change the slope of the observed quality-compute relationship. The practical difficulty is therefore not fitting a scaling curve once the experiments are done, but learning the forward relationship from model configuration to performance efficiently enough to decide which experiment to run next and ultimately solve the inverse design problem: selecting the best configuration to deploy under a scaling budget.

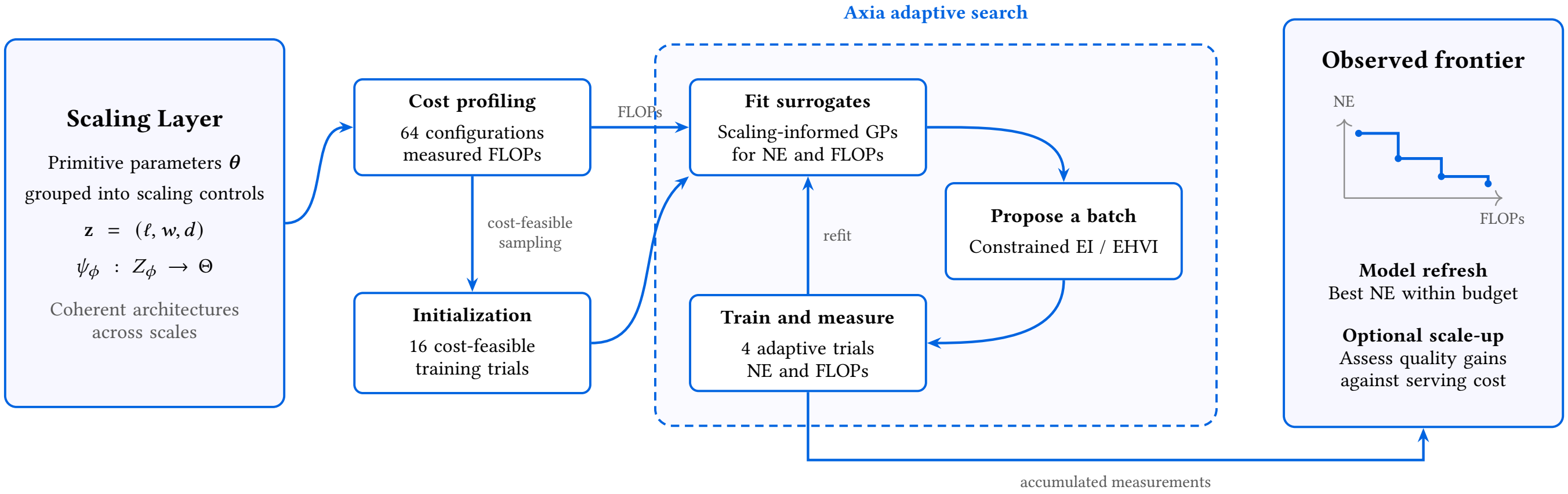
\begin{figure*}[t]
\centering
\resizebox{\textwidth}{!}{%
\begin{tikzpicture}[
  panel/.style={rectangle, rounded corners=4pt, draw=metablue,
    line width=0.8pt, fill=blue!3, text width=3.0cm,
    align=center, inner sep=7pt, minimum height=4.6cm,
    font=\footnotesize},
  stage/.style={rectangle, rounded corners=3pt, draw=metablue,
    line width=0.8pt, fill=white, text width=2.6cm,
    align=center, inner sep=5pt, minimum height=1.2cm,
    font=\footnotesize},
  engine/.style={rectangle, rounded corners=6pt, draw=metablue,
    line width=0.8pt, fill=blue!2, dash pattern=on 3pt off 2pt},
  flow/.style={-{Stealth[length=5pt,width=4pt]}, draw=metablue,
    line width=0.9pt},
  lbl/.style={font=\scriptsize, text=black!65, align=center},
]

\node[panel] (SL) at (0,0) {
  \textbf{\normalsize Scaling Layer}\\[7pt]
  Primitive parameters $\boldsymbol{\theta}$\\[3pt]
  grouped into scaling controls\\[5pt]
  $\mathbf{z}=(\ell,w,d)$\\[5pt]
  $\psi_\phi:Z_\phi\!\to\!\Theta$\\[7pt]
  {\color{black!65}Coherent architectures\\across scales}
};

\node[stage] (prof) at (4.1,1.2) {
  \textbf{Cost profiling}\\[3pt]
  64 configurations\\measured FLOPs
};
\node[stage] (init) at (4.1,-1.5) {
  \textbf{Initialization}\\[3pt]
  16 cost-feasible\\training trials
};
\draw[flow] (SL.east) to[out=0,in=180] (prof.west);
\draw[flow] (prof.south) -- node[lbl,right=2pt] {cost-feasible\\sampling} (init.north);

\node[stage] (fit) at (8.3,1.2) {
  \textbf{Fit surrogates}\\[3pt]
  Scaling-informed GPs\\for NE and FLOPs
};
\node[stage] (acq) at (11.5,-0.1) {
  \textbf{Propose a batch}\\[3pt]
  Constrained EI / EHVI
};
\node[stage] (train) at (8.3,-1.5) {
  \textbf{Train and measure}\\[3pt]
  4 adaptive trials\\NE and FLOPs
};
\draw[flow] (prof.east) -- node[lbl,above] {FLOPs} (fit.west);
\draw[flow] (init.east) to[out=0,in=220] (fit.south west);
\draw[flow] (fit.east) to[out=0,in=90] (acq.north);
\draw[flow] (acq.south) to[out=-90,in=0] (train.east);
\draw[flow] (train.north) -- node[lbl,right=2pt] {refit} (fit.south);
\begin{scope}[on background layer]
  \node[engine, fit=(fit)(acq)(train), inner sep=12pt] (ENG) {};
\end{scope}
\node[font=\footnotesize\bfseries, text=metablue, above=4pt of ENG.north]
  {Axia adaptive search};

\node[panel, minimum height=5.1cm] (OUT) at (16,0) {};
\node[font=\normalsize\bfseries, anchor=north, align=center]
  at ([yshift=-8pt]OUT.north) {Observed frontier};
\begin{scope}[shift={($(OUT.north)+(0,-1.8)$)}, scale=0.9]
  \draw[->, black!45, line width=0.5pt] (-1.1,-0.5) -- (1.1,-0.5)
    node[below=2pt, font=\scriptsize, black!65] {FLOPs};
  \draw[->, black!45, line width=0.5pt] (-1.1,-0.5) -- (-1.1,0.6)
    node[above=1pt, font=\scriptsize, black!65] {NE};
  \draw[metablue, line width=1pt]
    (-0.9,0.4) -- (-0.35,0.4) -- (-0.35,0.05)
    -- (0.25,0.05) -- (0.25,-0.2) -- (0.9,-0.2) -- (0.9,-0.3);
  \foreach \x/\y in {-0.9/0.4,-0.35/0.05,0.25/-0.2,0.9/-0.3}
    \fill[metablue] (\x,\y) circle (1.4pt);
\end{scope}
\node[font=\footnotesize, text width=2.9cm, align=center, anchor=south]
  at ([yshift=10pt]OUT.south) {
    \textbf{Model refresh}\\
    Best NE within budget\\[7pt]
    \textbf{Optional scale-up}\\
    Assess quality gains\\against serving cost
  };
\draw[flow] (train.south) -- ++(0,-0.85) -|
  node[lbl,pos=0.25,below=2pt] {accumulated measurements} (OUT.south);

\end{tikzpicture}%
}
\caption{The workflow at a glance. The Scaling Layer maps a compact vector of architectural controls $\mathbf{z}=(\ell,w,d)$ to valid model configurations. Cost profiling guides a cost-feasible initialization batch; scaling-informed quality and cost surrogates then guide adaptive training through constrained EI or EHVI. The main studies use 16 initialization and four adaptive trials; further batches repeat the propose--train--refit loop. The observed NE--COST(inference complexity measured in FLOPs) frontier supports a model refresh within an allowance or a scale-up decision based on the return from additional serving compute.}
\label{fig:axia-system}
\ifdefined\Description
\Description{A Scaling Layer feeds cost profiling and cost-feasible initialization. Cost and training observations inform scaling-informed surrogates in an adaptive loop that proposes a batch, trains its architectures, and refits. Accumulated measured NE and FLOPs form an observed frontier for choosing a model refresh or optional scale-up.}
\fi
\end{figure*}

We turn scaling laws into an automated decision-making mechanism. We introduce a \textbf{Scaling Layer} that exposes coupled architectural parameters through a consistent set of scaling controls, such as embedding dimension, interaction depth, and interaction width, which map into complete, valid configurations of an existing model template. This reduces a large collection of interdependent parameters to a compact and interpretable search space in which models can be instantiated coherently at any target scale. On top of that interface, we introduce \textbf{Axia}, an uncertainty-aware system that turns learned scaling relationships into budget-aware experimentation and deployment decisions. Axia combines analytical cost profiling, targeted survey experiments, scaling-law estimation, Pareto-aware candidate generation, and full-fidelity validation into a single closed loop (Figure~\ref{fig:axia-system}). It maintains a probabilistic model of the performance--cost surface, selects configurations that are both promising and informative based on the serving budget and an acquisition function chosen based on business use-case, and updates its estimate after each batch.

Deployed across 34 model iterations over two development cycles on Meta's ads ranking and retrieval fleet, the workflow delivered a mean offline normalized-entropy improvement of 0.66\% and a cumulative key metric online lift of 0.36\%. These results demonstrate the practical value of learning model-specific performance--cost relationships and using them to guide scaling decisions across a production ranking fleet. 

\section{Related Work}

\paragraph{Scaling Laws for Recommendation and Ranking.} Empirical scaling laws relate model quality to capacity, data, and computation~\citep{hestness2017deep}, and in large language models have proved regular enough to support extrapolation and compute allocation~\citep{kaplan2020scaling, hoffmann2022training}. Fitted conclusions are sensitive to the functional family and to which experiments are selected for fitting~\citep{li2025misfitting}, which motivates treating experimental design as part of scaling-law estimation; cheaper paired observations can substitute for expensive ones~\citep{brandfonbrener2025losstoloss}.

Analogous regularities hold in recommendation. Ardalani et al.~\citep{ardalani2022understanding} fit normalized-entropy-versus-resource frontiers with $L(x)=\alpha x^{-\beta}+\gamma$, where $x$ denotes data size, parameter count, or training compute, and $\alpha,\beta,\gamma$ are fitted constants. They find different compute efficiencies for embedding and MLP scaling strategies. DHEN reports diminishing quality gains with interaction depth and improvements over widening an MoE at comparable training compute~\citep{zhang2022dhen}. Wukong demonstrated predictable improvement when scaling the depth and width of high-order interaction models~\citep{zhang2024wukong}, HSTU reported power-law improvement with training compute for generative sequential recommendation~\citep{zhai2024actions}, and LLaTTE studied predictable sequence-model scaling~\citep{xiong2026llatte}, while Guan et al.~\citep{guan2026makeitlong} scaled user-history to lengths of $10,000$ in production, reporting monotonic gains. These studies establish that architectural choices affect scaling efficiency and provide strategies for allocating capacity. Our focus is learning that allocation per model with a small adaptive training budget, across coupled axes under a serving constraint.

Several studies use fitted scaling behavior for model-design decisions. P-Law predicts ranking metrics from depth, embedding dimension, and data scale, and uses the fit to identify configurations under model-size constraints~\citep{shen2025plaw}. Wang et al.~\citep{wang2024scalinglaws} estimate online scaling for advertisement retrieval and apply it to ROI-constrained design. Axia closes an additional loop: it represents uncertainty in the learned relationship and uses it to decide which costly configuration to evaluate next, resulting in model-specific scaling law, applicable across a heterogeneous fleet.

Work on hyperparameter scaling in language models expresses optimal training configurations as functions of a single scalar notion of size, whether empirically~\citep{mccandlish2018largebatch, kaplan2020scaling} or through parameterizations permitting zero-shot transfer across width~\citep{yang2021tensorv}; extending such transfer to additional capacity axes requires separate treatment~\citep{yang2024tensorvi}. Our scaling layer deliberately exposes several axes at once, which is why we revisit the question (Appendix~\ref{app:training_params}).
\vspace{-10pt}

\paragraph{Recommendation Model Architectures.} Modern ranking models combine large embedding tables, interaction modules, sequence encoders, and task-specific heads -- DLRM~\citep{naumov2019dlrm}, DHEN~\citep{zhang2022dhen}, and InterFormer~\citep{zeng2025interformer} -- which increase expressiveness but introduce coupled dimensions of capacity that must be allocated under serving constraints. These dimensions cannot be scaled independently: increasing embedding dimension can produce diminishing returns or embedding collapse~\citep{guo2024embeddingcollapse} and MixFormer~\citep{huang2026mixformer} studies the allocation between dense interaction and sequence modeling under a fixed budget through a unified architecture. Axia addresses the same allocation problem from a complementary direction, learning how to distribute capacity across the scaling dimensions of an existing template rather than prescribing a new architecture.

\paragraph{Bayesian Optimization and Experimental Design.} Bayesian optimization selects expensive evaluations using a probabilistic surrogate and an acquisition function~\citep{snoek2012practical}. High-dimensional methods exploit low effective dimension through random embeddings~\citep{wang2016rembo,letham2020alebo}, sparse axis-aligned priors~\citep{eriksson2021saasbo}, or searches over subsets of coordinates~\citep{li2017dropout}. Axia instead constructs its search coordinates from architectural semantics: each control moves a coherent group of primitive parameters together. This makes the parameterization itself a design choice that we evaluate alongside the search procedure.

Multi-fidelity BO allocates evaluations across cheaper approximations to a target objective~\citep{klein2017fast,wu2020practical}. Axia's profiling stage exploits a complementary asymmetry: inference complexity can be measured before training supplies a quality observation. These measurements guide cost-feasible initialization and inform subsequent constrained search. Batch, noise-aware expected improvement and expected hypervolume improvement provide established criteria for selecting evaluations~\citep{ament2023unexpected,daulton2021parallel}. We combine these criteria with scaling-informed surrogates to support two deployment decisions: improving quality within an allowance and exploring the complexity--quality frontier before committing to a scale-up.

Recent work also makes scaling-law estimation adaptive. \textit{Spend Less, Fit Better} treats scaling-law fitting as budget-aware experimental design~\citep{li2026spend}; \textit{Power-Law Entropy Search} selects experiments to reduce uncertainty in optimal-hyperparameter scaling laws per unit computational cost~\citep{chen2026efficiently}. These methods target accurate scaling-law estimates. Our search targets architectures that improve quality under serving constraints or expand the attained frontier. Axia brings together a ranking-specific parameterization, preliminary cost profiling, and scaling-informed adaptive search to make these decisions with a small number of parallel training batches.

\section{Scaling Layer}
\label{sec:param}

Large Web platforms deploy ranking and retrieval across many distinct surfaces, each with its own data distribution, optimization objective, and response-time requirement. This results in hundreds of unique models that often share a common template~\citep{liu2026template} (Figure~\ref{fig:parameterization}), but differ in signals, objectives, and serving capacity — and therefore in optimal configuration. For each model, we have limited capacity to explore offline before moving to online testing. Our objective is thus to maximize performance beneath each model's serving capacity while exhausting a limited number of offline trials.

Let $\boldsymbol{\theta} \in \Theta$ denote the primitive architecture parameters of a model template: embedding dimensions, interaction depths, hidden layer widths, attention dimensions, and sequence-processing capacities. Direct optimization over $\Theta$ is impractical given its high dimensionality and the limited number of training runs available.

A scaling layer is a pair $\phi = (G,h)$, where $G = \{G_1, \ldots, G_k\}$ partitions the primitive parameters into $k$ semantically related groups. The scaling rules $h = (h_1,\ldots, h_k)$ assign to each group a map $h_j: Z_j \rightarrow \Theta_{G_j}$ taking a single transformation to every parameter in that group. Together they induce a $k$-dimensional hyper-hyperparameter (HHP) space $Z = \{Z_1\times\ldots\times Z_k\}$ and a parameterization

\begin{equation}
    \psi: Z \rightarrow \Theta, \quad \psi(\mathbf{z})|_{G_j}=h_j(z_j)
\end{equation}

Some coordinates are naturally discrete -- interaction depth: $\ell \in \{1, 2, \ldots, N\}$ -- while other continuous parameters like interaction width are bounded on intervals and discretized at a resolution matching the snapping granularity of the underlying primitive parameters. The low dimensionality of $Z$ is what renders the resulting grid searchable: a space-filling design of a few dozen points covers it densely enough to support frontier estimation, which would be infeasible at the dimensionality of $\Theta$.

A point $\mathbf{z} \in Z$ is a vector of HHPs; for the scaling layer we arrive at Section~\ref{sec:current_scaling_layer}, $k=3$ and $\mathbf{z}=(\ell,w,d)$ which denote interaction depth, interaction width, and embedding dimension. Every parameter in a group is thus a function of that group's single coordinate, which is what makes the architecture resizable without becoming mis-proportioned. The architectures reachable under $\phi$ are exactly $\psi(Z)$.
Unlike conventional neural architecture search, which seeks an optimal $\boldsymbol{\theta}$, our objective is to optimize $\phi$ itself: we seek a scaling layer whose reachable set consistently attains frontiers that dominate those of alternative scaling layers under a matched run budget.

\subsection{Design Principles}
\label{sec:design_principles}
We derive a practical template scaling layer by iteratively applying a set of design principles motivated by search efficiency, transferability, and Pareto coverage: (1) \textbf{Capacity Coherence.} Parameters that jointly determine a common notion of model capacity should scale together; (2)  \textbf{Independent Performance Axes.} Parameters known to induce qualitatively different behavior should remain independently controllable; (3)  \textbf{Searchability.} The dimensionality of the induced HHP space should remain small enough to permit reliable frontier estimation under realistic search budgets; (4) \textbf{Transferability.} Groups should correspond to architectural concepts expected to generalize across tasks.

\subsection{Current Scaling Layer}
\label{sec:current_scaling_layer}
\begin{figure}[h!]
    \centering
    \includegraphics[height=0.5\textheight, keepaspectratio]{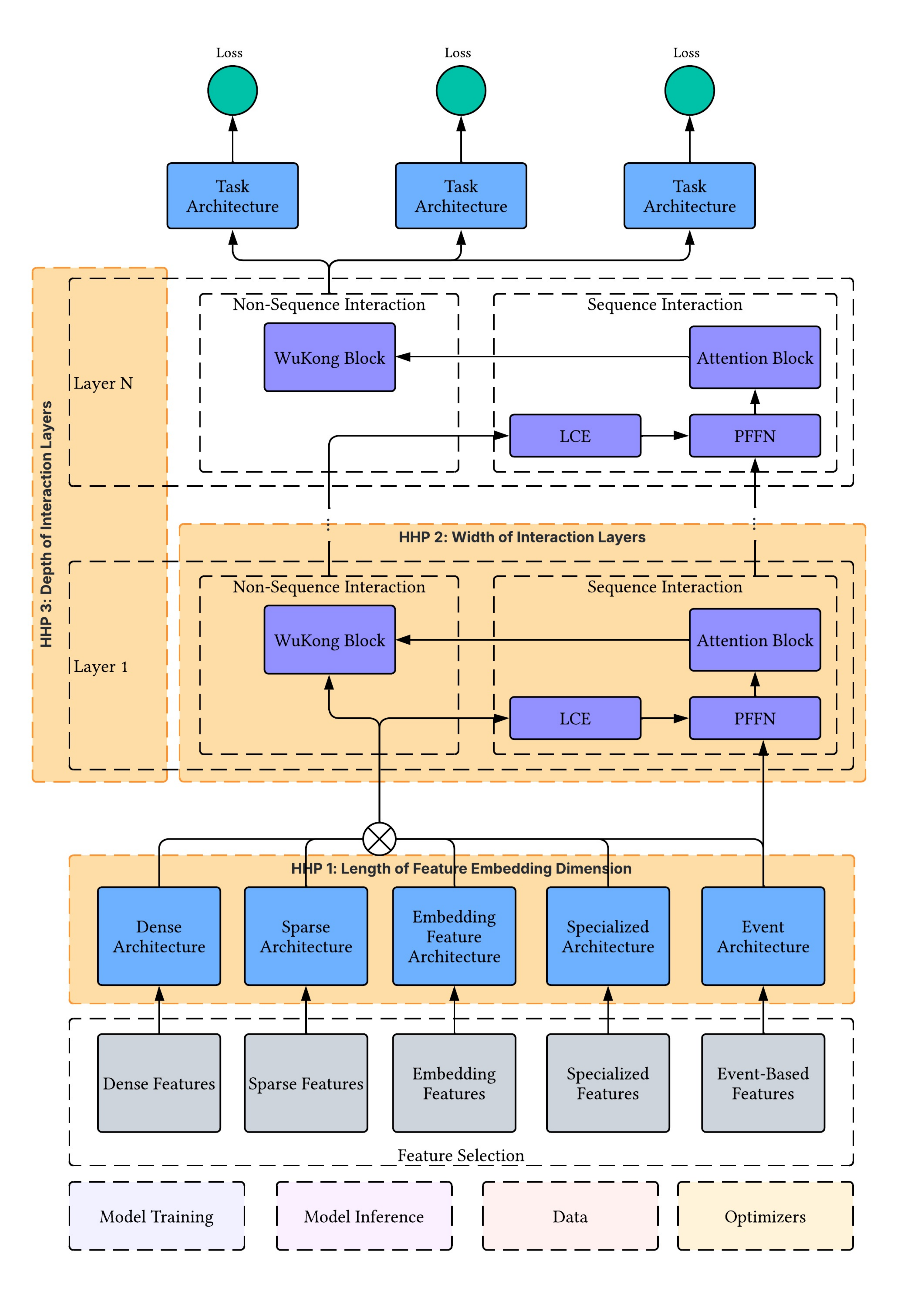}
    \caption{Primitive architecture parameters $\boldsymbol{\theta}$ are grouped into three semantically meaningful scaling axes: embedding dimension $d$, interaction depth $\ell$, and interaction width $w$, an umbrella parameter that jointly controls dimensions of multiple components in the interaction architecture.}
    \label{fig:parameterization}
\Description{Architecture diagram showing feature encoders feeding repeated sequence and non-sequence interaction layers, followed by task-specific outputs. Highlighted regions identify three scaling axes: embedding dimension, interaction width, and interaction depth.}
\end{figure}

Applying these principles under our run budget yields a scaling layer with $k=3$ coordinates: embedding dimension $d$, interaction depth $\ell$, and interaction width $w$. The first two are near-atomic -- each governs essentially one primitive parameter, controlling feature representation capacity and the order of feature interaction respectively. The third is an umbrella coordinate: $w$ jointly scales the hidden dimensions of multiple components within the interaction architecture, which the design principles group together because they determine a common notion of interaction capacity and are not independently useful to control at this budget. Figure~\ref{fig:parameterization} shows the grouping applied over a template.

Neither the cardinality nor the rules follow from the principles alone. The principles constrain $\phi$ -- they rule out groupings that split a coherent capacity notion, or that collapse axes known to behave differently -- but they do not pick a unique $(G, h)$, and the admissible choices are not enumerable at the cost of a full training run per evaluation. Both components were therefore settled empirically, under the matched-budget protocol of Section~\ref{sec:eval}, and we report those studies alongside our other results in Section~\ref{sec:results}. We emphasize that the layer presented here is the current outcome of that procedure rather than a fixed artifact: because Section~\ref{sec:eval} prices any candidate layer in the same units, the scaling layer remains an object of study, and refinements to it are adjudicated by the same evaluation that adjudicates changes to the engine. 
\section{Scaling Engine}
\label{sec:scaling}

Given a scaling layer $\phi$ (Section~\ref{sec:param}), the scaling engine searches the induced HHP space $Z_\phi$. For an instantiated architecture $\boldsymbol{\theta}=\psi_\phi(\mathbf{z})$, we write $\NE(\mathbf{z})$ and $\COST(\mathbf{z})>0$ for its normalized entropy and serving cost, e.g. inference complexity measured in FLOPs. The search objective depends on the deployment decision. Let $B$ be the target complexity and $\COST_0$ the complexity of the incumbent model. For a \emph{model refresh}, we seek better quality at or below the incumbent's complexity:
\begin{equation}
\label{eq:true_problem}
\mathbf{z}^* = \arg\min_{\mathbf{z}\in Z_\phi}\NE(\mathbf{z})
\quad\text{s.t.}\quad \COST(\mathbf{z})\leq B.
\end{equation}
For a scale-up target $B>\COST_0$, however, the team need not commit to deploying at that complexity. The target bounds exploration: the search should reveal the complexity--quality frontier up to $B$, allowing the team to assess quality gains relative to the incumbent and diminishing returns from additional compute before choosing what to ship. Each model ranks for a particular Web surface with its own latency and capacity constraints, so this operating-point decision is made per model rather than once for the fleet.

The attainable frontier depends on both the scaling layer and how it is searched. A compact, structured scaling layer makes the problem tractable, but each quality observation still requires training an architecture. Axia combines cost profiling, scaling-informed response models, and uncertainty-aware batch selection to spend this limited training budget. The acquisition functions are established Bayesian optimization methods; the system couples them to the architectural scaling layer and a staged workflow in which cost can be profiled before quality is observed.

\subsection{Scaling-Informed Surrogates}
\label{sec:joint_surrogate}

Wukong's scaling study demonstrates sustained quality gains from multiplicative increases in dense compute, with different returns from scaling individual interaction components~\citep{zhang2024wukong}. Axia uses scaling-informed response models to discover the best attainable quality at each serving budget, retaining architectural variation and predictive uncertainty. Multiplicative changes in architectural capacity motivate modeling selected coordinates on a logarithmic scale. For the production parameterization $\mathbf{z}=(\ell,w,d)$, the model used in our main comparison takes
\begin{equation}
\label{eq:input_transform}
\mathbf{u}=T(\mathbf{z})=(\log \ell,\ \log w,\ d),
\end{equation}
followed by normalization to the search bounds. Depth and width therefore express relative changes, while embedding dimension remains linear. This mixed transform is an empirical choice informed by our surrogate comparisons, allowing representation capacity and interaction capacity to have different response geometries. We model observed weighted NE on relevant prediction tasks on its original scale and Cost on a logarithmic scale, standardizing each response for fitting. The precise transformations and baseline settings are given in Appendix~\ref{app:experiment_details}.

A compact family that also permits sharing information between the outputs is
\begin{equation}
\label{eq:hierarchical}
\begin{aligned}
f_{\log\COST}(\mathbf{u})&=m_F+h_F(\mathbf{u}),\\
f_{\NE}(\mathbf{u})&=m_N+\beta h_F(\mathbf{u})+h_N(\mathbf{u}),
\end{aligned}
\qquad h_F\perp h_N,
\end{equation}
where $h_j\sim\mathcal{GP}(0,k_j)$ and $m_j$ are learned constants; affine input normalization and response standardization are suppressed in the notation. The reported experiments use $\beta=0$, corresponding to independent GPs for quality and cost. Each output has its own sum of an ARD RBF kernel and a linear kernel,
\begin{equation}
\label{eq:scaling_kernel}
k_j(\mathbf{u},\mathbf{u}')=\sigma_j^2 k_{\mathrm{RBF},j}(\mathbf{u},\mathbf{u}')+\mathbf{u}^\top\boldsymbol{\Lambda}_j \mathbf{u}',
\end{equation}
with nonnegative diagonal $\boldsymbol{\Lambda}_j$. The linear component represents a random linear trend in the transformed coordinates; the RBF component captures departures from that trend. When both the response and an input are log-transformed, the linear trend corresponds to a power law: $\log y=a+b\log x$ implies $y=e^a x^b$. The cost GP therefore supports power-law trends in depth and width, with RBF deviations. NE is a normalized cross-entropy~\citep{he2014practical}; we apply no additional logarithm to this response. Its linear component is therefore affine in log-transformed depth and width, with a linear dependence on embedding dimension.

These choices retain architecture-specific variation alongside scaling trends. Individual architectures can trade additional cost for worse NE, while the best attainable NE under an increasing COST allowance is necessarily non-increasing. Cost profiling supplies additional observations to the cost GP, whose posterior represents uncertainty between profiled configurations.

\subsection{Acquisition-Driven Candidate Selection}
\label{sec:acquisition}

After the initialization batch, Axia conditions the surrogates on observed quality and cost and selects a batch using their posterior distributions. This permits exploration beyond a frontier traced by point predictions. Our default deployment policy uses constrained expected improvement (EI) when $B\leq\COST_0$ and expected hypervolume improvement (EHVI) when $B>\COST_0$. The first supports a refresh within an allowance; the second supports deciding how much of a potential scale-up is worthwhile.

\paragraph{Constrained EI}
For a model refresh, constrained EI favors candidates that can improve weighted NE within the COST allowance. Posterior samples account for uncertainty in quality, cost feasibility, and the incumbent. This concentrates the training budget on improving the selected operating point.

\paragraph{Expected hypervolume improvement.}
When additional compute is available but its deployment is conditional on the return, we seek useful improvements across the complexity--quality frontier~\citep{daulton2021parallel}. The batch utility is
\begin{equation}
\alpha_{\mathrm{HV}}(Z_q)=
\mathbb{E}\!\left[\mathrm{HV}_{\mathbf{r}}(\Pareto_f(\mathcal D\cup Z_q))
-\mathrm{HV}_{\mathbf{r}}(\Pareto_f(\mathcal D))\right],
\end{equation}
where $Z_q$ is the proposed batch, $\mathcal D$ is the previously evaluated configurations, $\mathbf{r}$ is a fixed reference point, and $\Pareto_f$ denotes the feasible nondominated outcomes in a posterior sample. The resulting frontier supports a return-on-investment assessment of quality gains versus additional serving compute. Its reference point bounds the region whose improvement is rewarded. The team combines this frontier with the business value of quality and the cost of capacity to choose a deployment.

The rule based on incumbent complexity is an operational default. Constrained EI can also serve a larger target when the decision is simply to minimize NE under that allowance; frontier exploration can serve a smaller target when further compute savings matter. Both acquisitions use the scaling-informed posterior. We compare them under a common COST ceiling; both use batched, noise-aware, numerically stable implementations, detailed in Appendix~\ref{app:acquisition_details}.

\paragraph{Batching and adaptivity.}
At a fixed training budget, batch size determines how often the search can learn from completed evaluations. Smaller batches provide more opportunities to update the surrogate and redirect subsequent evaluations, but introduce additional sequential training rounds, potentially increasing the time to a final recommendation. Larger batches exploit parallel capacity but commit more evaluations before receiving feedback. We use 16 initialization trials followed by four adaptive trials, and compare alternative schedules to assess the benefit of additional adaptivity relative to the number of sequential rounds required (Section~\ref{sec:batching_schedules}).

Appendix~\ref{app:extensions} discusses extensions for sharing information across outputs and experiments, using training progression, and imposing multiple serving constraints.

\section{Evaluation Framework}
\label{sec:eval}
Because a change to the parameterization or the engine may shift the attainable cost--performance frontier, evaluating one requires rerunning each method under a matched budget and comparing observed frontiers against a fixed benchmark. We form a small representative model set $\mathcal{M}_R$ following the construction of Liu et al.~\citep{liu2026template} and hold it fixed, so that improvements to parameterization, surrogate modeling, and transfer are compared consistently. 

The output of a campaign for model $m$ is a set of evaluated configurations with true cost--performance pairs $(\COST(\mathbf{z}),\mathrm{NE}(\mathbf{z}))$ from full training, from which we extract the Pareto front $F_m$. Both objectives are minimized, so the front is summarized by its attainment function --- the best NE reached within a given cost:

\begin{equation}
\label{eq:attainment} 
P_{F_m}(\tilde{c}) = \min\bigl\{\mathrm{NE}(\mathbf{z}) : \mathbf{z} \in F_m,\, \COST(\mathbf{z}) \le \tilde{c}B_m\bigr\}
\end{equation} 

with cost expressed relative to $B_m$, model $m$'s serving budget prior to the campaign, so $\tilde{c}$ is a scale-up factor rather than an absolute cost. A change is scored by how much it improves attainment --- but attainment at which cost? Evaluating each model at its own current operating point measures improvement at the few budgets where the representative models happen to sit, making the criterion sensitive to their sizes rather than to the change under test; weighting all costs equally instead spends it on regions the fleet never operates in. We therefore weight attainment by the scale-ups the fleet actually undertakes: 

\begin{equation}
\label{eq:fleet_utility} 
\Delta \mathrm{FU} = \sum_{m\in\mathcal{M}_R} \alpha_m \int \bigl[P_{F_{m,0}}(\tilde{c}) - P_{F_{m,1}}(\tilde{c})\bigr]\, w(\tilde{c})\,d\tilde{c} 
\end{equation} 

Here $w$ is a density over relative cost with $\int w = 1$, estimated from realized scale-up factors across the fleet over a fixed historical window; it is prespecified and shared by all compared methods, since a weighting chosen after seeing results is a tunable parameter rather than a criterion. The integral is evaluated as a finite sum on the grid at which fronts are extracted. The weight $\alpha_m$ is a prespecified per-model importance summing to one --- we use revenue share, though any prespecified weighting serves. Because revenue share tracks the traffic each model serves, $\alpha_m$ weights the fleet by Web-surface exposure, so fleet utility measures improvement where the platform's Web traffic concentrates rather than uniformly across the representative set. Weighting by integration rather than ad-hoc averaging preserves Pareto compliance: for any non-negative $w$, a front that dominates another cannot score lower~\citep{zitzler2007thehypervolume}. Weighting over the cost axis alone leaves the NE direction unweighted, which gives the indicator NE-scale units; and because the difference of attainment functions is reference-free wherever both fronts lie below a common NE level, no per-model reference point is required. 

Two special cases are reported as sensitivity: uniform $w$ recovers the unweighted hypervolume difference~\citep{zitzler1999multiobjective}, which flags regression outside the fleet's operating range, and $w$ concentrated at a single scale-up recovers attainment at one operating point. Where method ordering is stable across the three weightings we report that stability; where it is not, we report the disagreement. 

Axia's surrogates are probabilistic, so we propagate uncertainty rather than assume it away: we draw from the surrogate posterior, re-extract the front per draw, and push predictive uncertainty through the Pareto operator to obtain interval estimates for $\Delta\mathrm{FU}$. Where model-level effects are aggregated we additionally apply a prespecified nonparametric bootstrap over independent model-level effects, and report all indicators under a matched trial budget. 

A change is adopted when both conditions hold: aggregate $\Delta\mathrm{FU}$ exceeds the detectability threshold of 0.03\%, below which it is not distinguishable from measurement noise, and the lower bound of its 95\% interval is non-negative. Either alone is insufficient --- the first admits intervals spanning zero, the second admits effects too small to matter. These thresholds document the internal decision rule governing the deployments of Section~\ref{sec:results}.

\section{Results}
\label{sec:results}
We evaluate the workflow on Meta's ads ranking and retrieval fleet over two completed cycles; a third is in progress and is not reported. The workflow delivered a mean relative offline NE improvement of 0.66\% over the deployment baselines it replaced, with a cumulative online lift of 0.36\%. The evaluation combines two tiers of evidence. 

\textbf{Ablations.} The first is controlled: holding the search space fixed, we compare scaling layers and candidate generation strategies under a matched budget of 20 full training runs, split into 16 initialization and four adaptive trials for adaptive approaches. We demonstrate these results on a single representative model, which admits direct inspection of the attained frontier. These comparisons provide a detailed view of search behavior; adoption decisions were made on fleet utility across the full representative model set. 

\textbf{Deployment Comparison.} The second tier is observational. Across 34 model iterations, each campaign's selected operating point replaced the incumbent configuration of one production model, measured against that deployment baseline -- the configuration produced by the manually orchestrated per-model process our approach supersedes. These observational outcomes summarize realized gains across completed, shipped campaigns, using each deployment baseline as the comparator.

\textbf{Measurements.} Offline model quality is Normalized Entropy (NE)~\citep{he2014practical}, the cross-entropy loss normalized by the entropy of the training data's background event rate; a per-model improvement of $\geq 0.05\%$ is conventionally treated as material, which is distinct from the aggregate $\Delta FU$ threshold of Section~\ref{sec:eval}. Online model quality is measured by the primary business metrics measured during live-traffic A/B tests.

\subsection{Choosing the Scaling Layer}
The design principles of Section~\ref{sec:design_principles} constrain $\phi$ but do not determine it. Two questions remain, one per component: how many groups should $G$ contain, and what rules $h$ should govern each? Neither admits an a priori answer, because both trade against a fixed budget of full training runs. 

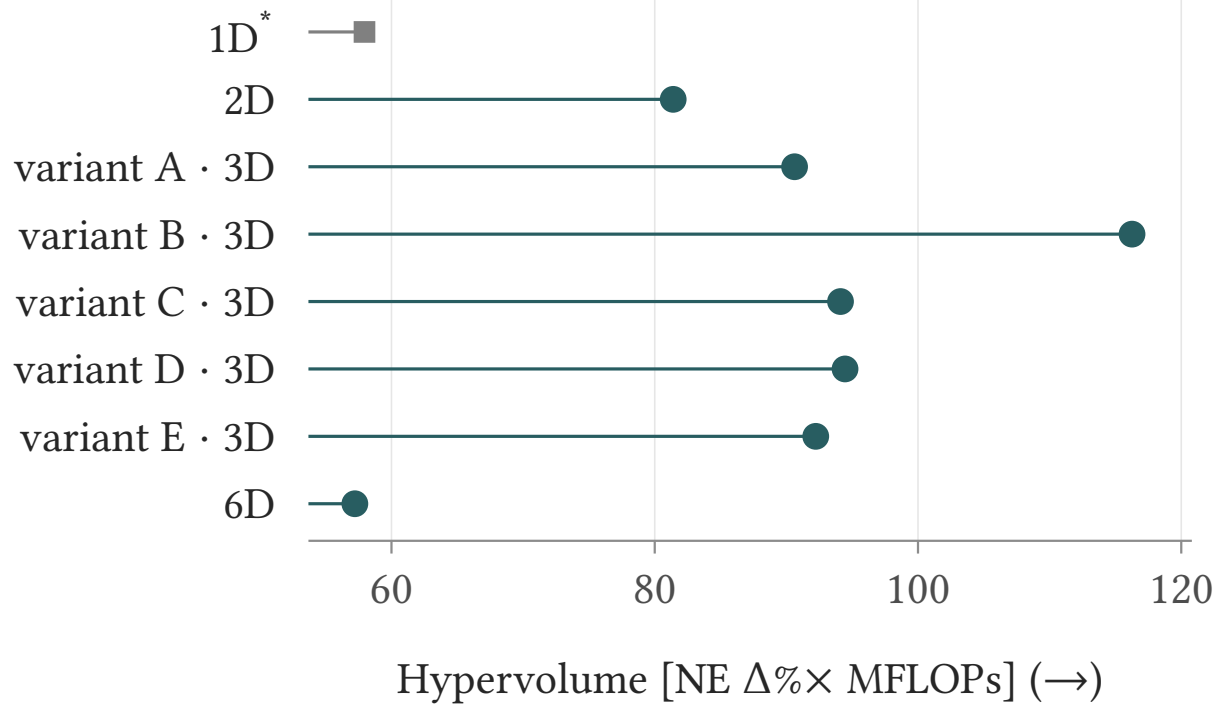
\begin{figure}[t]
    \centering
    \resizebox{\linewidth}{!}{\maxsizebox{\linewidth}{0.25\textheight}{\begingroup
\definecolor{ink}{HTML}{285D61}
\pgfplotsset{keep/.style 2 args={x filter/.append code={
\edef\actual{\thisrow{#1}}\def\expected{#2}
\ifx\actual\expected\else\def\pgfmathresult{nan}\fi}}}
\begin{tikzpicture}
\pgfmathsetlengthmacro{\summaryplotwidth}{\linewidth-3.6cm}
\pgfmathsetlengthmacro{\summaryplotheight}{\summaryplotwidth/1.61803398875}
\begin{axis}[width=\summaryplotwidth,height=\summaryplotheight,scale only axis,
font=\footnotesize,
axis x line*=bottom,axis y line*=left,
x axis line style={black!45,line width=.35pt},y axis line style={draw=none},
tick align=outside,xtick style={black!45,line width=.35pt},ytick style={draw=none},
major tick length=2pt,
xmajorgrids=true,major grid style={black!10,line width=.25pt},
xmin=53.691275167785,xmax=120.805369127517,xtick={60,80,100,120},
scaled x ticks=false,xticklabel style={/pgf/number format/fixed,/pgf/number format/precision=2,text=black!70},
ymin=-.55,ymax=7.55,y dir=reverse,ytick={0,1,2,3,4,5,6,7},
yticklabels={1D\textsuperscript{*},2D,variant A $\cdot$ 3D,variant B $\cdot$ 3D,variant C $\cdot$ 3D,variant D $\cdot$ 3D,variant E $\cdot$ 3D,6D},
yticklabel style={text=black!85},
xlabel={Hypervolume [NE $\Delta\%\times$ MFLOPs] ($\rightarrow$)},
xlabel style={yshift=-3pt,text=black!85},clip=false]
\addplot[only marks,unbounded coords=discard,mark=square*,mark size=1.5pt,gray,keep={metric}{final_hypervolume_relative_ne},keep={method}{brute_force},keep={status}{complete},error bars/.cd,x dir=minus,x explicit,error mark=none,error bar style={line width=.5pt},/pgfplots/.cd] table[col sep=comma,x=value,y=row,x error expr=(\thisrow{value}+0)-(53.691275167785)] {figures/parameterization-summary/results-long.csv};
\addplot[only marks,unbounded coords=discard,mark=*,mark size=1.9pt,ink,keep={metric}{final_hypervolume_relative_ne},keep={method}{embedding_linear_qlognei},keep={status}{complete},error bars/.cd,x dir=minus,x explicit,error mark=none,error bar style={line width=.5pt},/pgfplots/.cd] table[col sep=comma,x=value,y=row,x error expr=(\thisrow{value}+0)-(53.691275167785)] {figures/parameterization-summary/results-long.csv};
\end{axis}
\end{tikzpicture}
\endgroup}}
    \caption{Final attained hypervolume across scaling layers, using 16 initial and four adaptive trials under a 160 MFLOPs cap. Hypervolume is measured in relative NE--MFLOPs space against a common reference point; higher is better. 3-dimensional variants are defined in Table~\ref{tab:scaling_layer_variants}. \textsuperscript{*}1D uses a 20-point 1D grid.}
    \label{fig:parameterization-hypervolume}
    \ifdefined\Description
        \Description{Final attained hypervolume in relative NE percent times MFLOPs. Teal circles show AxiaV2 searches; the gray square is the 1D grid reference. Higher values are better.}
    \fi
\end{figure}

\begin{figure}[t]
    \centering
    \input{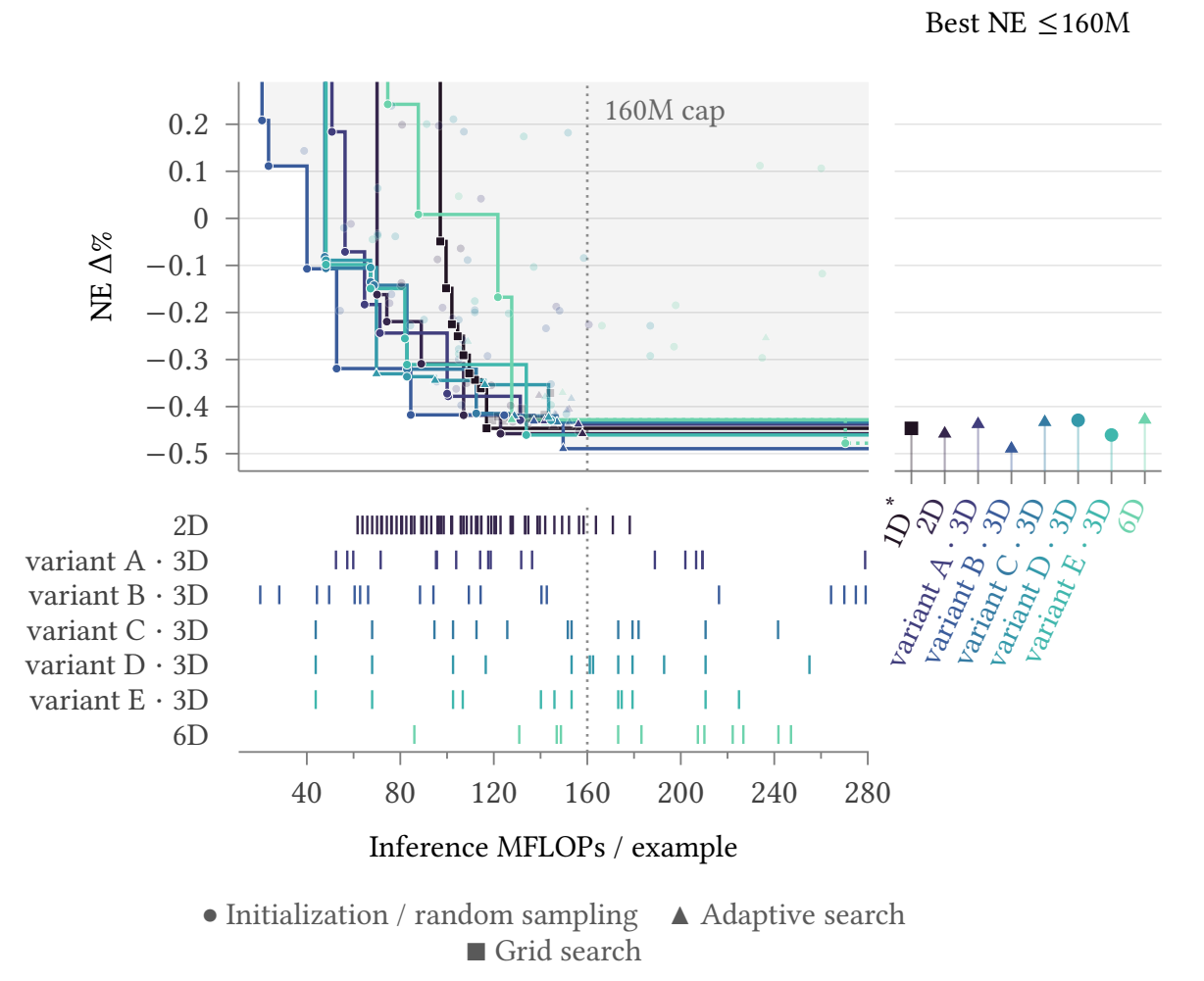}
    \caption{Observed NE--COST(MFLOPs) frontiers across scaling layers with 16 initialization and four adaptive trials. NE shows relative change from a shared baseline; lower is better. Solid lines show feasible frontiers and best-at-160M plateaus; dotted lines show efficient above-cap results. 3-dimensional variants are defined in Table~\ref{tab:scaling_layer_variants}. Faint shading denotes pooled dominance. Bottom rugs show cost profiles; the right panel shows best feasible NE. \textsuperscript{*}1D uses a 20-point 1D grid.}
    \label{fig:parameterization-frontiers}
    \ifdefined\Description
        \Description{Observed relative NE versus inference MFLOPs for eight parameterizations, with a 160 MFLOPs cap. Colored step frontiers distinguish parameterizations; circles show initialization, triangles adaptive search, and squares grid search.}
    \fi
\end{figure}

\subsubsection{Choosing the Number of Coordinates}
The number of groups $k$ sets the dimension of $Z$, and therefore how much of the space a fixed survey can cover. Figure~\ref{fig:parameterization-hypervolume} compares scaling layers at $k \in \{1, 2, 3, 6\}$ under the above specified budget of 16 initial and four adaptive samples. We omit $k\in\{4,5\}$ because our template's interaction stack decomposes coherently into three or six groups but not into four or five, and splitting a capacity group arbitrarily would violate the capacity-coherence principle of Section~\ref{sec:design_principles}. Improvement in hypervolume is observed from 1 to 2 to 3 coordinates and is lower at six. For this model and budget, the production three-coordinate parameterization attains the strongest frontier. More sample-efficient search could make additional coordinates useful at the same training budget.

The result reflects a tradeoff between search-space dimension and the observations available to explore it. A survey budget $n$ buys a given quality of frontier estimate over a space of dimension $k$; gains in sample efficiency raise the quality attainable at fixed $(n,k)$, and the surplus may be spent in either direction. Spent on $n$, it lowers cost per campaign -- dominant when iterating many models at once. Spent on $k$, it admits a coordinate previously unlearnable at this budget, extending the law to a capacity dimension formerly pinned by hand.

\subsubsection{Choosing the Scaling Rules}
At fixed $k$, the choices of $G$ and $h$ are under-determined: the design principles of Section~\ref{sec:design_principles} admit many assignments of primitive parameters to a group, and many monotone rules for moving them, and the space is not enumerable at the cost of a full training run per evaluation. We therefore proceed by hypothesis rather than by search, forming candidate refinements from how the interaction stack is known to consume capacity, and evaluating each under the matched-budget protocol of Section~\ref{sec:eval}. We tested the following hypotheses, with evaluation results shown in Figures~\ref{fig:parameterization-hypervolume} and ~\ref{fig:parameterization-frontiers}:

\begin{table}[th]
\centering
\small
\begin{tabular}{@{}llll@{}}
\toprule
\textbf{Variant} & \textbf{Coverage} & \textbf{Proportionality} & \textbf{Profile} \\
\midrule
A & non-sequence & uniform & flat \\
B & sequence + non-sequence & uniform & flat \\
C & sequence + non-sequence & capacity & flat \\
D & sequence + non-sequence & capacity & waist \\
E & sequence + non-sequence & capacity & taper \\
\bottomrule
\end{tabular}
\caption{Scaling layer variants. Each 3-dimensional variant modifies the interaction-width coordinate $w$ along one factor: coverage (which primitives $w$ owns within the model interaction), proportionality (uniform vs. capacity-weighted scaling), and profile (flat, waist, or taper shape across interaction layers). Coverage changes the grouping $G$; proportionality and profile change the rules $h$ (Section~\ref{sec:param}).}
\label{tab:scaling_layer_variants}
\end{table}

These studies yield the scaling layer (Figure~\ref{fig:parameterization}) defined in Section~\ref{sec:current_scaling_layer} and used in the subsequent search-method ablations. The sensitivity of the frontier to $h$ at fixed $k$ is itself a finding: the number of coordinates is not sufficient to characterize a scaling layer, and two layers of equal dimension can differ materially in attainable frontier. This is why the protocol of Section~\ref{sec:eval} adjudicates changes to $h$ on the same footing as changes to $k$.

\subsection{Candidate Quality at Matched Budget}
\label{sec:candidate_quality}
Holding the scaling layer of Section~\ref{sec:current_scaling_layer} fixed, we compare six families of candidate-generation strategies, including two grids (Table~\ref{tab:search_strategies}). The non-grid methods have 20 completed training runs; adaptive methods use 16 initialization and four follow-up trials. The two grids are complete at 20 and 27 trials, respectively; the larger grid uses a higher training budget. Appendix~\ref{app:experiment_details} records the implemented settings and differences in initialization.

\begin{table}[th]
\centering
\small
\begin{tabular}{@{}lll@{}}
\toprule
\textbf{Strategy} & \textbf{Initialization} & \textbf{Follow-up} \\
\midrule
Grid Search (3$\times$3$\times$3) & Cartesian grid & None \\
Grid Search (2$\times$2$\times$5) & Cartesian grid & None \\
Random (Sobol) & 20 Sobol trials & None \\
BO (Ax) & 16 Sobol trials & Constrained EI \\
BO (Ax w/ cost profiling) & 16 cost-feasible & Default BO \\
Open-Loop Surrogate & 16 cost-feasible & Point predictions \\
Axia & 16 cost-feasible & Constrained EI \\
\bottomrule
\end{tabular}
\caption{Candidate-generation procedures. Profiling-enabled methods share the cost survey, but not their initialization training observations. Ax uses the shipped default surrogate and acquisition at the evaluated internal version; Axia uses the scaling-informed surrogate in Section~\ref{sec:joint_surrogate}.}
\label{tab:search_strategies}
\end{table}

\begin{figure}[t]
    \centering
    \input{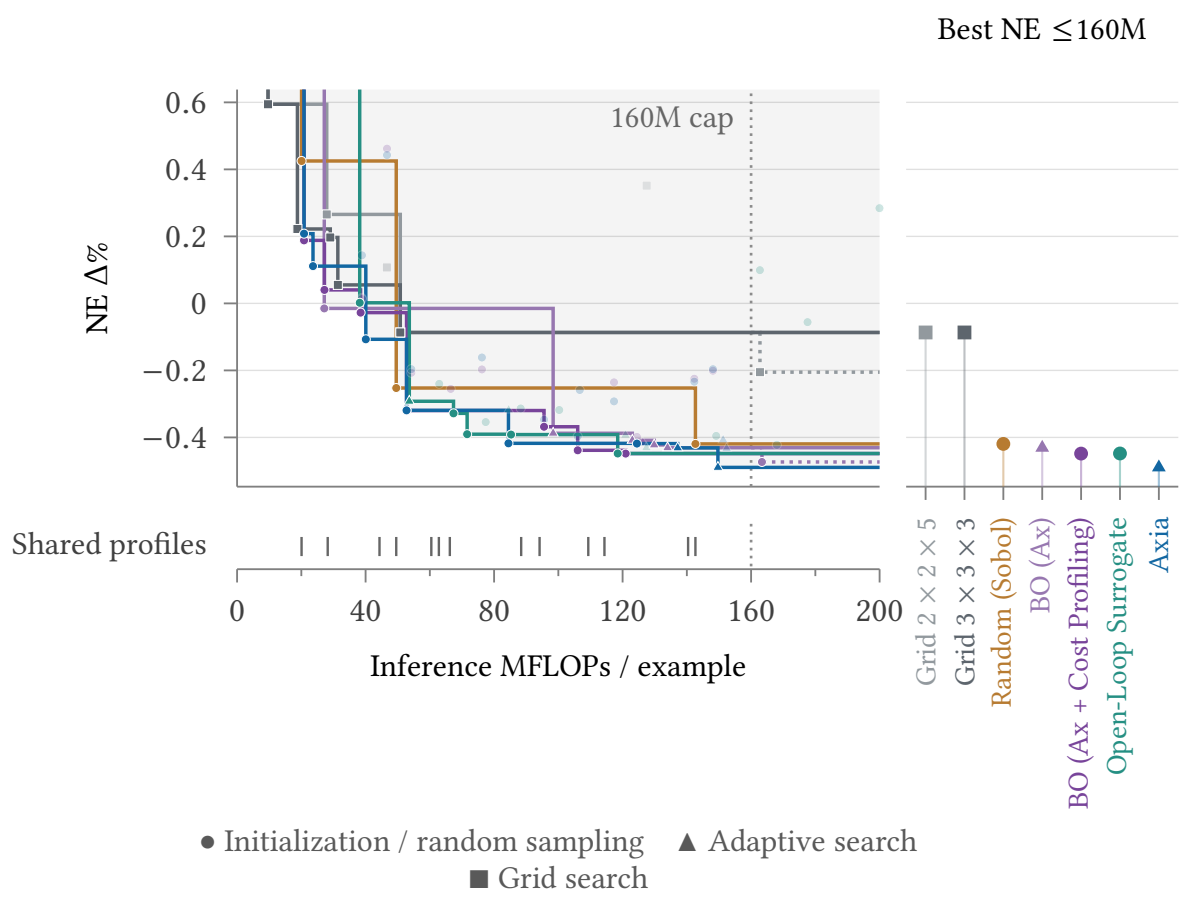}
    \caption{Observed NE--FLOPs frontiers across search methods using the strongest observed scaling layer. Adaptive methods use 16 initialization and four adaptive trials; Random (Sobol) uses 20 samples. NE is relative to a shared baseline; lower is better. Solid lines show frontiers feasible at 160 MFLOPs and their plateaus; dotted lines show efficient above-cap results. Faint shading denotes pooled dominance. The bottom rug shows 64 shared cost profiles used by BO (Ax w/ cost profiling), Open-Loop Surrogate, and Axia; the right panel shows best feasible NE.}
    \label{fig:search-method-frontiers}
    \ifdefined\Description
        \Description{Seven search methods shown with colored step frontiers and matching labels on the right. Circles represent initialization or random sampling, triangles adaptive search, and squares grid search.}
    \fi
\end{figure}

Figure~\ref{fig:search-method-frontiers} shows attained frontiers for the representative model, using measured complexity--quality pairs. We report best feasible NE at 160 MFLOPs and hypervolume under a common reference, rather than a surrogate's predicted frontier.

Axia achieves the lowest feasible NE in this comparison: relative NE reductions of 0.070\%, 0.059\%, and 0.042\% against Random, BO (Ax~\citep{olson2025ax}), and Open-Loop Surrogate, respectively. Its attained hypervolume is 24.5\%, 19.0\%, and 7.9\% higher. Against BO (Ax + Cost Profiling), the relative NE reduction is 0.041\% and the hypervolume gain is 1.0\%.

\subsection{Batching Search Evaluations}
\label{sec:batching_schedules}

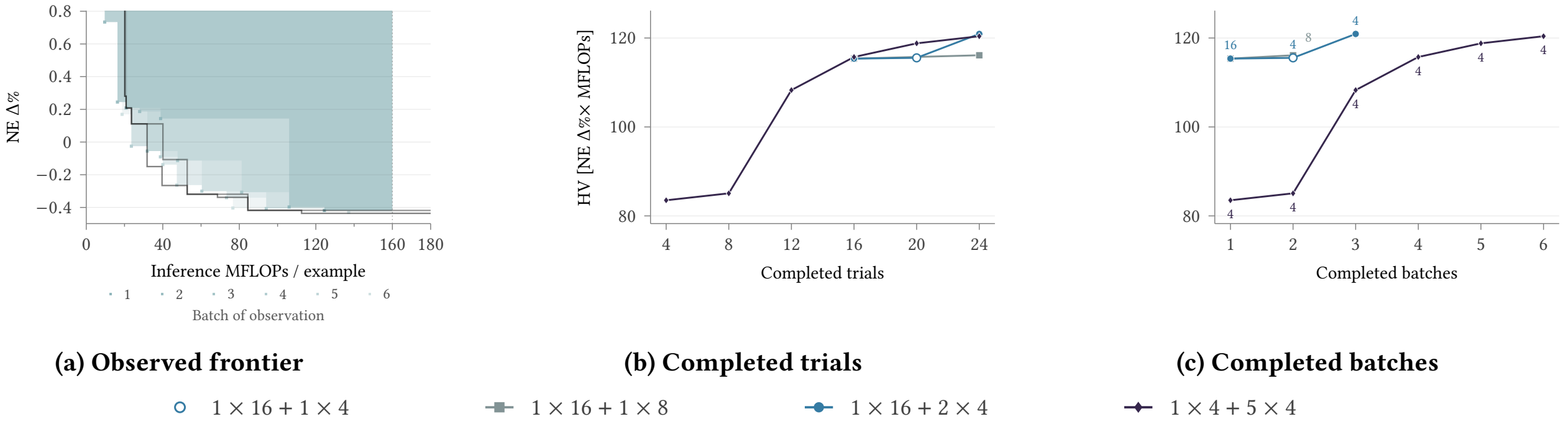
\begin{figure*}[t]
\centering
\begin{subfigure}[t]{.325\textwidth}
\centering
\resizebox{\linewidth}{!}{\begingroup

\pgfplotsset{compat=1.17}
\usepgfplotslibrary{fillbetween}
\pgfplotsset{keep/.style 2 args={x filter/.append code={
\edef\actual{\thisrow{#1}}\def\expected{#2}
\ifx\actual\expected\else\def\pgfmathresult{nan}\fi}}}
\pgfplotsset{paperaxis/.style={scale only axis,font=\fontsize{7}{8}\selectfont,
axis x line*=bottom,axis y line*=left,
axis line style={black!50,line width=.35pt},tick align=outside,
tick style={black!50,line width=.35pt},major tick length=2.4pt,minor tick length=1.2pt,
scaled ticks=false,tick label style={text=black!75},
ymajorgrids=true,major grid style={black!10,line width=.23pt},
clip=true,unbounded coords=discard,title style={font=\fontsize{7.5}{9}\selectfont,text=black!85},
xlabel style={font=\fontsize{7}{8}\selectfont},ylabel style={font=\fontsize{7}{8}\selectfont},
legend style={draw=none,font=\fontsize{6.5}{7.5}\selectfont,cells={anchor=west},column sep=6pt},
}}
\definecolor{region}{HTML}{357b83}
\definecolor{continuation}{HTML}{357ba3}
\definecolor{eight}{HTML}{829395}
\definecolor{six}{HTML}{36274d}

\definecolor{b1}{HTML}{91b7bc}
\definecolor{b2}{HTML}{9cbec2}
\definecolor{b3}{HTML}{a7c6c9}
\definecolor{b4}{HTML}{b4ced1}
\definecolor{b5}{HTML}{c2d7d9}
\definecolor{b6}{HTML}{d1e1e2}
\begin{tikzpicture}
\begin{axis}[paperaxis,set layers=standard,width=4.65cm,height=2.873858047687011cm,name=frontier,at={(0,0)},anchor=south west,xmin=0,xmax=180,ymin=-0.4986330424556886,ymax=0.8037369403906549,xtick={0,40,80,120,160},extra x ticks={180},minor x tick num=1,ytick distance=.2,yticklabel style={/pgf/number format/fixed,/pgf/number format/precision=2},xlabel={Inference MFLOPs / example},xlabel style={at={(axis description cs:0.5,-0.23)},anchor=center},ylabel={NE $\Delta\%$}]
\addplot[draw=none,fill=region,fill opacity=0.08,forget plot,keep={path_id}{b1_fill}] table[col sep=comma,x=mflops,y=ne_delta_percent] {figures/batch-schedules/prod3d-qlognehvi-schedules/paths.csv} -- cycle;
\addplot[draw=none,fill=region,fill opacity=0.08,forget plot,keep={path_id}{b2_fill}] table[col sep=comma,x=mflops,y=ne_delta_percent] {figures/batch-schedules/prod3d-qlognehvi-schedules/paths.csv} -- cycle;
\addplot[draw=none,fill=region,fill opacity=0.08,forget plot,keep={path_id}{b3_fill}] table[col sep=comma,x=mflops,y=ne_delta_percent] {figures/batch-schedules/prod3d-qlognehvi-schedules/paths.csv} -- cycle;
\addplot[draw=none,fill=region,fill opacity=0.08,forget plot,keep={path_id}{b4_fill}] table[col sep=comma,x=mflops,y=ne_delta_percent] {figures/batch-schedules/prod3d-qlognehvi-schedules/paths.csv} -- cycle;
\addplot[draw=none,fill=region,fill opacity=0.08,forget plot,keep={path_id}{b5_fill}] table[col sep=comma,x=mflops,y=ne_delta_percent] {figures/batch-schedules/prod3d-qlognehvi-schedules/paths.csv} -- cycle;
\addplot[draw=none,fill=region,fill opacity=0.08,forget plot,keep={path_id}{b6_fill}] table[col sep=comma,x=mflops,y=ne_delta_percent] {figures/batch-schedules/prod3d-qlognehvi-schedules/paths.csv} -- cycle;
\draw[black!40,densely dotted,line width=.35pt] (axis cs:160,-0.4986330424556886) -- (axis cs:160,0.8037369403906549);
\addplot[only marks,mark=square*,mark size=.4pt,b1,mark options={draw=none,fill=b1},forget plot,keep={observed_batch}{1},keep={draw_marker}{1}] table[col sep=comma,x=mflops,y=ne_delta_percent] {figures/batch-schedules/prod3d-qlognehvi-schedules/observations.csv};
\addplot[only marks,mark=square*,mark size=.4pt,b2,mark options={draw=none,fill=b2},forget plot,keep={observed_batch}{2},keep={draw_marker}{1}] table[col sep=comma,x=mflops,y=ne_delta_percent] {figures/batch-schedules/prod3d-qlognehvi-schedules/observations.csv};
\addplot[only marks,mark=square*,mark size=.4pt,b3,mark options={draw=none,fill=b3},forget plot,keep={observed_batch}{3},keep={draw_marker}{1}] table[col sep=comma,x=mflops,y=ne_delta_percent] {figures/batch-schedules/prod3d-qlognehvi-schedules/observations.csv};
\addplot[only marks,mark=square*,mark size=.4pt,b4,mark options={draw=none,fill=b4},forget plot,keep={observed_batch}{4},keep={draw_marker}{1}] table[col sep=comma,x=mflops,y=ne_delta_percent] {figures/batch-schedules/prod3d-qlognehvi-schedules/observations.csv};
\addplot[only marks,mark=square*,mark size=.4pt,b5,mark options={draw=none,fill=b5},forget plot,keep={observed_batch}{5},keep={draw_marker}{1}] table[col sep=comma,x=mflops,y=ne_delta_percent] {figures/batch-schedules/prod3d-qlognehvi-schedules/observations.csv};
\addplot[only marks,mark=square*,mark size=.4pt,b6,mark options={draw=none,fill=b6},forget plot,keep={observed_batch}{6},keep={draw_marker}{1}] table[col sep=comma,x=mflops,y=ne_delta_percent] {figures/batch-schedules/prod3d-qlognehvi-schedules/observations.csv};
\addplot[black,solid,on layer=axis foreground,draw opacity=0.15,line width=.55pt,no marks,forget plot,keep={path_id}{ref1_line}] table[col sep=comma,x=mflops,y=ne_delta_percent] {figures/batch-schedules/prod3d-qlognehvi-schedules/paths.csv};
\addplot[black,solid,on layer=axis foreground,draw opacity=0.3,line width=.55pt,no marks,forget plot,keep={path_id}{ref2_line}] table[col sep=comma,x=mflops,y=ne_delta_percent] {figures/batch-schedules/prod3d-qlognehvi-schedules/paths.csv};
\addplot[black,solid,on layer=axis foreground,draw opacity=0.5,line width=.55pt,no marks,forget plot,keep={path_id}{ref3_line}] table[col sep=comma,x=mflops,y=ne_delta_percent] {figures/batch-schedules/prod3d-qlognehvi-schedules/paths.csv};
\end{axis}
\path[fill=b1,draw=none] (0.3250000000000005cm,-.95cm) ++(-.4pt,-.4pt) rectangle ++(.8pt,.8pt);\node[anchor=west,font=\fontsize{6}{7}\selectfont,text=black!70] at (0.3950000000000005cm,-.95cm) {1};
\path[fill=b2,draw=none] (1.0250000000000006cm,-.95cm) ++(-.4pt,-.4pt) rectangle ++(.8pt,.8pt);\node[anchor=west,font=\fontsize{6}{7}\selectfont,text=black!70] at (1.0950000000000006cm,-.95cm) {2};
\path[fill=b3,draw=none] (1.7250000000000005cm,-.95cm) ++(-.4pt,-.4pt) rectangle ++(.8pt,.8pt);\node[anchor=west,font=\fontsize{6}{7}\selectfont,text=black!70] at (1.7950000000000006cm,-.95cm) {3};
\path[fill=b4,draw=none] (2.4250000000000003cm,-.95cm) ++(-.4pt,-.4pt) rectangle ++(.8pt,.8pt);\node[anchor=west,font=\fontsize{6}{7}\selectfont,text=black!70] at (2.495cm,-.95cm) {4};
\path[fill=b5,draw=none] (3.1250000000000004cm,-.95cm) ++(-.4pt,-.4pt) rectangle ++(.8pt,.8pt);\node[anchor=west,font=\fontsize{6}{7}\selectfont,text=black!70] at (3.1950000000000003cm,-.95cm) {5};
\path[fill=b6,draw=none] (3.8250000000000006cm,-.95cm) ++(-.4pt,-.4pt) rectangle ++(.8pt,.8pt);\node[anchor=west,font=\fontsize{6}{7}\selectfont,text=black!70] at (3.8950000000000005cm,-.95cm) {6};
\node[anchor=center,font=\fontsize{6}{7}\selectfont,text=black!65] at (2.325cm,-1.23cm) {Batch of observation};
\pgfresetboundingbox
\path[use as bounding box] (-1.30cm,-1.40cm) rectangle (4.85cm,3.123858047687011cm);
\end{tikzpicture}
\endgroup}
\caption{Observed frontier}
\end{subfigure}
\hfill
\begin{subfigure}[t]{.325\textwidth}
\centering
\resizebox{\linewidth}{!}{\begingroup

\pgfplotsset{compat=1.17}
\usepgfplotslibrary{fillbetween}
\pgfplotsset{keep/.style 2 args={x filter/.append code={
\edef\actual{\thisrow{#1}}\def\expected{#2}
\ifx\actual\expected\else\def\pgfmathresult{nan}\fi}}}
\pgfplotsset{paperaxis/.style={scale only axis,font=\fontsize{7}{8}\selectfont,
axis x line*=bottom,axis y line*=left,
axis line style={black!50,line width=.35pt},tick align=outside,
tick style={black!50,line width=.35pt},major tick length=2.4pt,minor tick length=1.2pt,
scaled ticks=false,tick label style={text=black!75},
ymajorgrids=true,major grid style={black!10,line width=.23pt},
clip=true,unbounded coords=discard,title style={font=\fontsize{7.5}{9}\selectfont,text=black!85},
xlabel style={font=\fontsize{7}{8}\selectfont},ylabel style={font=\fontsize{7}{8}\selectfont},
legend style={draw=none,font=\fontsize{6.5}{7.5}\selectfont,cells={anchor=west},column sep=6pt},
}}
\definecolor{region}{HTML}{357b83}
\definecolor{continuation}{HTML}{357ba3}
\definecolor{eight}{HTML}{829395}
\definecolor{six}{HTML}{36274d}

\definecolor{b1}{HTML}{91b7bc}
\definecolor{b2}{HTML}{9cbec2}
\definecolor{b3}{HTML}{a7c6c9}
\definecolor{b4}{HTML}{b4ced1}
\definecolor{b5}{HTML}{c2d7d9}
\definecolor{b6}{HTML}{d1e1e2}
\begin{tikzpicture}
\begin{axis}[paperaxis,width=4.65cm,height=2.873858047687011cm,name=b,at={(0,0)},anchor=south west,xmin=3,xmax=25,ymin=78.2793146437826,ymax=126.15659816378603,xtick={4,8,12,16,20,24},yticklabel style={/pgf/number format/fixed,/pgf/number format/precision=0},xlabel={Completed trials},ylabel={HV [NE $\Delta\%\times$ MFLOPs]}]
\addplot[eight,solid,line width=.7pt,mark=square*,mark size=1.35pt,mark options={draw=white,line width=.2pt},forget plot,keep={schedule}{eight},keep={metric}{hv}] table[col sep=comma,x=trials,y=value] {figures/batch-schedules/prod3d-qlognehvi-schedules/trajectories.csv};

\addplot[continuation,solid,line width=.7pt,mark=*,mark size=1.35pt,mark options={draw=white,line width=.2pt},forget plot,keep={schedule}{continuation},keep={metric}{hv}] table[col sep=comma,x=trials,y=value] {figures/batch-schedules/prod3d-qlognehvi-schedules/trajectories.csv};

\addplot[six,solid,line width=.7pt,mark=diamond*,mark size=1.35pt,mark options={draw=white,line width=.2pt},forget plot,keep={schedule}{six},keep={metric}{hv}] table[col sep=comma,x=trials,y=value] {figures/batch-schedules/prod3d-qlognehvi-schedules/trajectories.csv};

\addplot[only marks,mark=*,mark size=1.5pt,continuation,mark options={fill=white,line width=.55pt},forget plot,keep={schedule}{two},keep={metric}{hv},keep={batch}{2}] table[col sep=comma,x=trials,y=value] {figures/batch-schedules/prod3d-qlognehvi-schedules/trajectories.csv};

\end{axis}
\pgfresetboundingbox
\path[use as bounding box] (-1.30cm,-1.40cm) rectangle (4.85cm,3.123858047687011cm);
\end{tikzpicture}
\endgroup}
\caption{Completed trials}
\end{subfigure}
\hfill
\begin{subfigure}[t]{.325\textwidth}
\centering
\resizebox{\linewidth}{!}{\begingroup

\pgfplotsset{compat=1.17}
\usepgfplotslibrary{fillbetween}
\pgfplotsset{keep/.style 2 args={x filter/.append code={
\edef\actual{\thisrow{#1}}\def\expected{#2}
\ifx\actual\expected\else\def\pgfmathresult{nan}\fi}}}
\pgfplotsset{paperaxis/.style={scale only axis,font=\fontsize{7}{8}\selectfont,
axis x line*=bottom,axis y line*=left,
axis line style={black!50,line width=.35pt},tick align=outside,
tick style={black!50,line width=.35pt},major tick length=2.4pt,minor tick length=1.2pt,
scaled ticks=false,tick label style={text=black!75},
ymajorgrids=true,major grid style={black!10,line width=.23pt},
clip=true,unbounded coords=discard,title style={font=\fontsize{7.5}{9}\selectfont,text=black!85},
xlabel style={font=\fontsize{7}{8}\selectfont},ylabel style={font=\fontsize{7}{8}\selectfont},
legend style={draw=none,font=\fontsize{6.5}{7.5}\selectfont,cells={anchor=west},column sep=6pt},
}}
\definecolor{region}{HTML}{357b83}
\definecolor{continuation}{HTML}{357ba3}
\definecolor{eight}{HTML}{829395}
\definecolor{six}{HTML}{36274d}

\definecolor{b1}{HTML}{91b7bc}
\definecolor{b2}{HTML}{9cbec2}
\definecolor{b3}{HTML}{a7c6c9}
\definecolor{b4}{HTML}{b4ced1}
\definecolor{b5}{HTML}{c2d7d9}
\definecolor{b6}{HTML}{d1e1e2}
\begin{tikzpicture}
\begin{axis}[paperaxis,width=4.65cm,height=2.873858047687011cm,name=c,at={(0,0)},anchor=south west,xmin=0.75,xmax=6.25,ymin=78.2793146437826,ymax=126.15659816378603,xtick={1,2,3,4,5,6},yticklabel style={/pgf/number format/fixed,/pgf/number format/precision=0},xlabel={Completed batches},ylabel={}]
\addplot[eight,solid,line width=.7pt,mark=square*,mark size=1.35pt,mark options={draw=white,line width=.2pt},forget plot,keep={schedule}{eight},keep={metric}{hv}] table[col sep=comma,x=batch,y=value] {figures/batch-schedules/prod3d-qlognehvi-schedules/trajectories.csv};

\addplot[draw=none,only marks,mark=none,forget plot,keep={schedule}{eight},keep={metric}{hv},keep={annotate}{1},point meta=explicit,nodes near coords,every node near coord/.append style={font=\fontsize{5}{6}\selectfont,text=eight,anchor=south west,xshift=1.5pt,yshift=2pt}] table[col sep=comma,x=batch,y=value,meta expr={\thisrow{batch} == 1 ? 16 : 8}] {figures/batch-schedules/prod3d-qlognehvi-schedules/trajectories.csv};

\addplot[continuation,solid,line width=.7pt,mark=*,mark size=1.35pt,mark options={draw=white,line width=.2pt},forget plot,keep={schedule}{continuation},keep={metric}{hv}] table[col sep=comma,x=batch,y=value] {figures/batch-schedules/prod3d-qlognehvi-schedules/trajectories.csv};

\addplot[draw=none,only marks,mark=none,forget plot,keep={schedule}{continuation},keep={metric}{hv},keep={annotate}{1},point meta=explicit,nodes near coords,every node near coord/.append style={font=\fontsize{5}{6}\selectfont,text=continuation,anchor=south,yshift=.35pt}] table[col sep=comma,x=batch,y=value,meta expr={\thisrow{batch} == 1 ? 16 : 4}] {figures/batch-schedules/prod3d-qlognehvi-schedules/trajectories.csv};

\addplot[six,solid,line width=.7pt,mark=diamond*,mark size=1.35pt,mark options={draw=white,line width=.2pt},forget plot,keep={schedule}{six},keep={metric}{hv}] table[col sep=comma,x=batch,y=value] {figures/batch-schedules/prod3d-qlognehvi-schedules/trajectories.csv};

\addplot[draw=none,only marks,mark=none,forget plot,keep={schedule}{six},keep={metric}{hv},point meta=explicit,nodes near coords,every node near coord/.append style={font=\fontsize{5}{6}\selectfont,text=six,anchor=north,yshift=-.35pt}] table[col sep=comma,x=batch,y=value,meta expr={4}] {figures/batch-schedules/prod3d-qlognehvi-schedules/trajectories.csv};

\addplot[only marks,mark=*,mark size=1.5pt,continuation,mark options={fill=white,line width=.55pt},forget plot,keep={schedule}{two},keep={metric}{hv},keep={batch}{2}] table[col sep=comma,x=batch,y=value] {figures/batch-schedules/prod3d-qlognehvi-schedules/trajectories.csv};

\end{axis}
\pgfresetboundingbox
\path[use as bounding box] (-1.30cm,-1.40cm) rectangle (4.85cm,3.123858047687011cm);
\end{tikzpicture}
\endgroup}
\caption{Completed batches}
\end{subfigure}
\par\smallskip
\begingroup
\definecolor{continuation}{HTML}{357BA3}
\definecolor{eight}{HTML}{829395}
\definecolor{six}{HTML}{36274D}
\begin{tikzpicture}
\draw[continuation,fill=white,line width=.55pt] (0.15,0) circle(1.5pt);
\node[anchor=west,text=black!75,font=\fontsize{7}{8}\selectfont] at (0.37,0) {$1\times16+1\times4$};
\draw[eight,solid,line width=.7pt] (3.4,0)--(3.6999999999999997,0);
\draw[eight] plot[mark=square*,mark size=1.35pt] coordinates {(3.55,0)};
\node[anchor=west,text=black!75,font=\fontsize{7}{8}\selectfont] at (3.77,0) {$1\times16+1\times8$};
\draw[continuation,solid,line width=.7pt] (6.8,0)--(7.1,0);
\draw[continuation] plot[mark=*,mark size=1.35pt] coordinates {(6.95,0)};
\node[anchor=west,text=black!75,font=\fontsize{7}{8}\selectfont] at (7.17,0) {$1\times16+2\times4$};
\draw[six,solid,line width=.7pt] (10.2,0)--(10.5,0);
\draw[six] plot[mark=diamond*,mark size=1.35pt] coordinates {(10.35,0)};
\node[anchor=west,text=black!75,font=\fontsize{7}{8}\selectfont] at (10.569999999999999,0) {$1\times4+5\times4$};
\end{tikzpicture}
\endgroup
\caption{Search progress under different batching schedules for the production parameterization: feasible hypervolume under a 160 MFLOPs cap. Notation is initialization batches $\times$ trials per batch, then adaptive batches $\times$ trials per batch. Solid black frontiers give the 16-trial initialization and its 20- and 24-trial continuations at increasing opacity, the 20-trial schedule appearing as a hollow checkpoint on the 24-trial curve. Nested translucent regions give the 4-trial-batch schedule, earlier-attained regions darker where layers overlap. Plotted points were Pareto-efficient after at least one batch; batch-axis numbers give trials per batch.}
\label{fig:batch-prod3d-qlognehvi-schedules-hypervolume}
\ifdefined\Description
\Description{Three panels show cumulative observed Pareto regions, progress versus completed trials, and progress versus completed batches. The first panel uses identical translucent fills and small square observations. The trajectories compare initialization and adaptive batch allocations under the same inference cost cap.}
\fi
\end{figure*}

Figure~\ref{fig:batch-prod3d-qlognehvi-schedules-hypervolume} compares batching schedules while holding the scaling layer and candidate-generation method fixed. Completed trials measure training expenditure; completed batches measure the sequential search rounds that contribute to turnaround time. At equal budgets, additional feedback rounds can help, with benefits depending on the surrogate and acquisition. The schedules expose the tradeoff between opportunities to adapt and the sequential training rounds needed for a final recommendation.

\subsection{Production Scaling Outcomes}
\label{sec:deployment_outcomes}
The practical test of a scaling workflow is whether it can reliably turn additional compute into better production outcomes. In our setting, infrastructure investments provide each model with a serving budget between $1.2\times$ and $10\times$ that of its production predecessor. The challenge is therefore not simply to build larger models, but to extract the greatest performance improvement from each new serving budget. Across two development cycles, our workflow was used to scale 34 production models (Table~\ref{tab:deployment_results}).

\paragraph{Offline.}
Across all 34 model iterations, the proposed workflow demonstrated consistency as normalized entropy (NE) improved by an average of 0.66\% relative to the corresponding deployment baselines (Table~\ref{tab:deployment_results}). Relative improvement is $100 \times \nicefrac{\mathrm{NE}_{\mathrm{base}}-\mathrm{NE}_{\mathrm{new}}}{\mathrm{NE}_{\mathrm{base}}}$.

Despite the difference in scale between the two cycles, 11 models in the first and 23 in the second, their mean improvements were nearly identical: 0.66\% and 0.67\%, respectively. This stability suggests that the results were not driven by an unusually favorable set of models in one cycle. Individual outcomes varied (overall SD $=0.59$ percentage points), as expected for a heterogeneous production fleet spanning different data distributions, objectives, and serving budgets.

\paragraph{Online.}
The critical test was whether the offline improvements would translate into measurable gains after deployment. Both development cycles produced positive online shifts (Table~\ref{tab:deployment_results}). Cycle~1 delivered a 0.07\% lift across 11 models, followed by a 0.29\% lift across 23 models in Cycle~2. 
At production scale, even incremental percentage improvements translate into substantial business impact, and these gains correspond to business impact on the order of hundreds of millions of dollars in annual revenue.

Taken together, the deployments establish a progression from additional serving compute, to consistent offline improvement, to measurable business impact. The reported cumulative lift of 0.36\% sums the two cycle-level shipped lifts over disjoint model sets (Table~\ref{tab:deployment_results}).

\begin{table}[th]
\centering
\small
\begin{tabular}{@{}lrrrr@{}}
\toprule
& & \multicolumn{2}{c}{\textbf{Offline}} & \textbf{Online} \\
\cmidrule(lr){3-4}
\textbf{Period} &
\textbf{Models} &
\textbf{Mean $\Delta$NE (\%)} &
\textbf{SD (pp)} &
\textbf{Lift (\%)} \\
\midrule
Cycle 1 & 11 & 0.66 & 0.49 & 0.07 \\
Cycle 2 & 23 & 0.67 & 0.66 & 0.29 \\
\midrule
\textbf{All deployments} &
\textbf{34} &
\textbf{0.66} &
\textbf{0.59} &
\textbf{$0.36^{\dagger}$} \\
\bottomrule
\end{tabular}
\caption{Offline and online outcomes across two development cycles.
Offline values in the final row are aggregate statistics over all 34
models. ${}^{\dagger}$The online value is the sum of the shipped
cycle-level lifts over disjoint model sets, not a separately measured
aggregate effect.}
\label{tab:deployment_results}
\end{table} 
\section{Conclusion}
Efficient scaling of ranking and retrieval models calls for the joint design of a template parameterization, an automated procedure for learning its complexity--quality frontier, and an evaluation protocol capable of adjudicating changes to either. Grouping coupled primitive parameters into a small set of hyper-hyperparameters makes a template resizable without becoming mis-proportioned; Axia then learns complexity--quality relationships over that space by acquisition-driven selection, returning an observed frontier from a small number of informative runs rather than an exhaustive sweep; and a fleet-utility criterion, computed on a fixed representative model set with uncertainty propagated through the Pareto operator, makes refinements to either component measurable.

Deployed at Meta across 34 model iterations spanning two development cycles, the system delivers an average NE gain of 0.66\% and a cumulative online lift of 0.36\% -- evidence of the workflow's practical use across a heterogeneous fleet. The contribution of this work is an automated and auditable system for learning complexity--quality relationships on arbitrary architectures -- one that replaces per-model, manually orchestrated studies with a reusable procedure, and thereby turns compute allocation across a large and continuously evolving model fleet into a systematic discipline.

\newpage
\bibliographystyle{ACM-Reference-Format} 
\bibliography{paper} 


\begin{thebibliography}{41}


\ifx \showCODEN    \undefined \def \showCODEN     #1{\unskip}     \fi
\ifx \showDOI      \undefined \def \showDOI       #1{#1}\fi
\ifx \showISBNx    \undefined \def \showISBNx     #1{\unskip}     \fi
\ifx \showISBNxiii \undefined \def \showISBNxiii  #1{\unskip}     \fi
\ifx \showISSN     \undefined \def \showISSN      #1{\unskip}     \fi
\ifx \showLCCN     \undefined \def \showLCCN      #1{\unskip}     \fi
\ifx \shownote     \undefined \def \shownote      #1{#1}          \fi
\ifx \showarticletitle \undefined \def \showarticletitle #1{#1}   \fi
\ifx \showURL      \undefined \def \showURL       {\relax}        \fi
\providecommand\bibfield[2]{#2}
\providecommand\bibinfo[2]{#2}
\providecommand\natexlab[1]{#1}
\providecommand\showeprint[2][]{arXiv:#2}

\bibitem[Ament et~al\mbox{.}(2023)]%
        {ament2023unexpected}
\bibfield{author}{\bibinfo{person}{Sebastian Ament}, \bibinfo{person}{Samuel
  Daulton}, \bibinfo{person}{David Eriksson}, \bibinfo{person}{Maximilian
  Balandat}, {and} \bibinfo{person}{Eytan Bakshy}.}
  \bibinfo{year}{2023}\natexlab{}.
\newblock \showarticletitle{Unexpected Improvements to Expected Improvement for
  {Bayesian} Optimization}. In \bibinfo{booktitle}{\emph{Advances in Neural
  Information Processing Systems}}, Vol.~\bibinfo{volume}{36}.
  \bibinfo{publisher}{Curran Associates, Inc.}, \bibinfo{pages}{20577--20612}.
\newblock
\urldef\tempurl%
\url{https://doi.org/10.52202/075280-0904}
\showDOI{\tempurl}


\bibitem[Ardalani et~al\mbox{.}(2022)]%
        {ardalani2022understanding}
\bibfield{author}{\bibinfo{person}{Newsha Ardalani},
  \bibinfo{person}{Carole-Jean Wu}, \bibinfo{person}{Zeliang Chen},
  \bibinfo{person}{Bhargav Bhushanam}, {and} \bibinfo{person}{Adnan Aziz}.}
  \bibinfo{year}{2022}\natexlab{}.
\newblock \bibinfo{title}{Understanding Scaling Laws for Recommendation
  Models}.
\newblock
\newblock
\showeprint[arxiv]{2208.08489}~[cs.IR]
\urldef\tempurl%
\url{https://arxiv.org/abs/2208.08489}
\showURL{%
\tempurl}


\bibitem[Brandfonbrener et~al\mbox{.}(2025)]%
        {brandfonbrener2025losstoloss}
\bibfield{author}{\bibinfo{person}{David Brandfonbrener},
  \bibinfo{person}{Nikhil Anand}, \bibinfo{person}{Nikhil Vyas},
  \bibinfo{person}{Eran Malach}, {and} \bibinfo{person}{Sham~M. Kakade}.}
  \bibinfo{year}{2025}\natexlab{}.
\newblock \showarticletitle{{Loss-to-Loss Prediction: Scaling Laws for All
  Datasets}}.
\newblock \bibinfo{journal}{\emph{Transactions on Machine Learning Research}}
  (\bibinfo{year}{2025}).
\newblock
\showeprint[arxiv]{2411.12925}
\urldef\tempurl%
\url{https://mlanthology.org/tmlr/2025/brandfonbrener2025tmlr-losstoloss/}
\showURL{%
\tempurl}


\bibitem[Chen et~al\mbox{.}(2026)]%
        {chen2026efficiently}
\bibfield{author}{\bibinfo{person}{Zhiliang Chen}, \bibinfo{person}{Sebastian
  Ament}, \bibinfo{person}{David Eriksson}, \bibinfo{person}{Maximilian
  Balandat}, \bibinfo{person}{Bryan Kian~Hsiang Low}, \bibinfo{person}{Eytan
  Bakshy}, {and} \bibinfo{person}{Jihao~Andreas Lin}.}
  \bibinfo{year}{2026}\natexlab{}.
\newblock \bibinfo{title}{Efficiently Estimating Optimal Hyperparameter Scaling
  Laws through Power-Law Entropy Search}.
\newblock
\newblock
\showeprint[arxiv]{2609.01431}~[cs.LG]
\urldef\tempurl%
\url{https://arxiv.org/abs/2609.01431}
\showURL{%
\tempurl}


\bibitem[Daulton et~al\mbox{.}(2021)]%
        {daulton2021parallel}
\bibfield{author}{\bibinfo{person}{Samuel Daulton}, \bibinfo{person}{Maximilian
  Balandat}, {and} \bibinfo{person}{Eytan Bakshy}.}
  \bibinfo{year}{2021}\natexlab{}.
\newblock \showarticletitle{Parallel {Bayesian} Optimization of Multiple Noisy
  Objectives with Expected Hypervolume Improvement}. In
  \bibinfo{booktitle}{\emph{Advances in Neural Information Processing
  Systems}}, Vol.~\bibinfo{volume}{34}. \bibinfo{publisher}{Curran Associates,
  Inc.}, \bibinfo{pages}{2187--2200}.
\newblock
\urldef\tempurl%
\url{https://proceedings.neurips.cc/paper/2021/hash/11704817e347269b7254e744b5e22dac-Abstract.html}
\showURL{%
\tempurl}


\bibitem[Domhan et~al\mbox{.}(2015)]%
        {domhan2015speeding}
\bibfield{author}{\bibinfo{person}{Tobias Domhan}, \bibinfo{person}{Jost~Tobias
  Springenberg}, {and} \bibinfo{person}{Frank Hutter}.}
  \bibinfo{year}{2015}\natexlab{}.
\newblock \showarticletitle{Speeding Up Automatic Hyperparameter Optimization
  of Deep Neural Networks by Extrapolation of Learning Curves}. In
  \bibinfo{booktitle}{\emph{Proceedings of the Twenty-Fourth International
  Joint Conference on Artificial Intelligence (IJCAI 2015)}}.
  \bibinfo{publisher}{AAAI Press}, \bibinfo{pages}{3460--3468}.
\newblock


\bibitem[Eriksson and Jankowiak(2021)]%
        {eriksson2021saasbo}
\bibfield{author}{\bibinfo{person}{David Eriksson} {and}
  \bibinfo{person}{Martin Jankowiak}.} \bibinfo{year}{2021}\natexlab{}.
\newblock \showarticletitle{High-dimensional {Bayesian} optimization with
  sparse axis-aligned subspaces}. In \bibinfo{booktitle}{\emph{Proceedings of
  the Thirty-Seventh Conference on Uncertainty in Artificial Intelligence}}
  \emph{(\bibinfo{series}{Proceedings of Machine Learning Research},
  Vol.~\bibinfo{volume}{161})}. \bibinfo{publisher}{PMLR},
  \bibinfo{pages}{493--503}.
\newblock
\urldef\tempurl%
\url{https://proceedings.mlr.press/v161/eriksson21a.html}
\showURL{%
\tempurl}


\bibitem[Falkner et~al\mbox{.}(2018)]%
        {falkner2018bohb}
\bibfield{author}{\bibinfo{person}{Stefan Falkner}, \bibinfo{person}{Aaron
  Klein}, {and} \bibinfo{person}{Frank Hutter}.}
  \bibinfo{year}{2018}\natexlab{}.
\newblock \showarticletitle{BOHB: Robust and Efficient Hyperparameter
  Optimization at Scale}. In \bibinfo{booktitle}{\emph{Proceedings of the 35th
  International Conference on Machine Learning (ICML)}}.
  \bibinfo{pages}{1436--1445}.
\newblock


\bibitem[Guan et~al\mbox{.}(2026)]%
        {guan2026makeitlong}
\bibfield{author}{\bibinfo{person}{Lin Guan}, \bibinfo{person}{Jia-Qi Yang},
  \bibinfo{person}{Zhishan Zhao}, \bibinfo{person}{Beichuan Zhang},
  \bibinfo{person}{Bo Sun}, \bibinfo{person}{Xuanyuan Luo}, {et~al\mbox{.}}}
  \bibinfo{year}{2026}\natexlab{}.
\newblock \showarticletitle{Make It Long, Keep It Fast: End-to-End 10k-Sequence
  Modeling at Billion Scale on Douyin Recommendation}. In
  \bibinfo{booktitle}{\emph{Proceedings of the ACM Web Conference 2026 (WWW
  '26)}}. \bibinfo{publisher}{ACM}, \bibinfo{address}{Dubai, United Arab
  Emirates}.
\newblock
\urldef\tempurl%
\url{https://doi.org/10.1145/3774904.3792811}
\showDOI{\tempurl}


\bibitem[Guo et~al\mbox{.}(2024)]%
        {guo2024embeddingcollapse}
\bibfield{author}{\bibinfo{person}{Xingzhuo Guo}, \bibinfo{person}{Junwei Pan},
  \bibinfo{person}{Ximei Wang}, \bibinfo{person}{Baixu Chen},
  \bibinfo{person}{Jie Jiang}, {and} \bibinfo{person}{Mingsheng Long}.}
  \bibinfo{year}{2024}\natexlab{}.
\newblock \showarticletitle{On the Embedding Collapse when Scaling up
  Recommendation Models}. In \bibinfo{booktitle}{\emph{Proceedings of the 41st
  International Conference on Machine Learning}}
  \emph{(\bibinfo{series}{Proceedings of Machine Learning Research},
  Vol.~\bibinfo{volume}{235})}. \bibinfo{publisher}{PMLR},
  \bibinfo{pages}{16891--16909}.
\newblock
\urldef\tempurl%
\url{https://proceedings.mlr.press/v235/guo24e.html}
\showURL{%
\tempurl}


\bibitem[He et~al\mbox{.}(2014)]%
        {he2014practical}
\bibfield{author}{\bibinfo{person}{Xinran He}, \bibinfo{person}{Junfeng Pan},
  \bibinfo{person}{Ou Jin}, \bibinfo{person}{Tianbing Xu}, \bibinfo{person}{Bo
  Liu}, \bibinfo{person}{Tao Xu}, \bibinfo{person}{Yanxin Shi},
  \bibinfo{person}{Antoine Atallah}, \bibinfo{person}{Ralf Herbrich},
  \bibinfo{person}{Stuart Bowers}, {et~al\mbox{.}}}
  \bibinfo{year}{2014}\natexlab{}.
\newblock \showarticletitle{Practical lessons from predicting clicks on ads at
  facebook}. In \bibinfo{booktitle}{\emph{Proceedings of the eighth
  international workshop on data mining for online advertising}}.
  \bibinfo{pages}{1--9}.
\newblock


\bibitem[Hestness et~al\mbox{.}(2017)]%
        {hestness2017deep}
\bibfield{author}{\bibinfo{person}{Joel Hestness}, \bibinfo{person}{Sharan
  Narang}, \bibinfo{person}{Newsha Ardalani}, \bibinfo{person}{Gregory Diamos},
  \bibinfo{person}{Heewoo Jun}, \bibinfo{person}{Hassan Kianinejad},
  \bibinfo{person}{Md. Mostofa~Ali Patwary}, \bibinfo{person}{Yang Yang}, {and}
  \bibinfo{person}{Yanqi Zhou}.} \bibinfo{year}{2017}\natexlab{}.
\newblock \bibinfo{title}{Deep Learning Scaling is Predictable, Empirically}.
\newblock
\newblock
\showeprint[arxiv]{1712.00409}~[cs.LG]
\urldef\tempurl%
\url{https://arxiv.org/abs/1712.00409}
\showURL{%
\tempurl}


\bibitem[Hoffmann et~al\mbox{.}(2022)]%
        {hoffmann2022training}
\bibfield{author}{\bibinfo{person}{Jordan Hoffmann}, \bibinfo{person}{Sebastian
  Borgeaud}, \bibinfo{person}{Arthur Mensch}, \bibinfo{person}{Elena
  Buchatskaya}, \bibinfo{person}{Trevor Cai}, \bibinfo{person}{Eliza
  Rutherford}, \bibinfo{person}{Diego de Las~Casas}, \bibinfo{person}{Lisa~Anne
  Hendricks}, \bibinfo{person}{Johannes Welbl}, \bibinfo{person}{Aidan Clark},
  \bibinfo{person}{Tom Hennigan}, \bibinfo{person}{Eric Noland},
  \bibinfo{person}{Katie Millican}, \bibinfo{person}{George van~den Driessche},
  \bibinfo{person}{Bogdan Damoc}, \bibinfo{person}{Aurelia Guy},
  \bibinfo{person}{Simon Osindero}, \bibinfo{person}{Karen Simonyan},
  \bibinfo{person}{Erich Elsen}, \bibinfo{person}{Jack~W. Rae},
  \bibinfo{person}{Oriol Vinyals}, {and} \bibinfo{person}{Laurent Sifre}.}
  \bibinfo{year}{2022}\natexlab{}.
\newblock \bibinfo{title}{Training Compute-Optimal Large Language Models}.
\newblock
\newblock
\showeprint[arxiv]{2203.15556}~[cs.CL]
\urldef\tempurl%
\url{https://arxiv.org/abs/2203.15556}
\showURL{%
\tempurl}


\bibitem[Huang et~al\mbox{.}(2026)]%
        {huang2026mixformer}
\bibfield{author}{\bibinfo{person}{Xu Huang}, \bibinfo{person}{Hao Zhang},
  \bibinfo{person}{Zhifang Fan}, \bibinfo{person}{Yunwen Huang},
  \bibinfo{person}{Zhuoxing Wei}, \bibinfo{person}{Zheng Chai},
  \bibinfo{person}{Jinan Ni}, \bibinfo{person}{Yuchao Zheng}, {and}
  \bibinfo{person}{Qiwei Chen}.} \bibinfo{year}{2026}\natexlab{}.
\newblock \showarticletitle{{MixFormer}: Co-Scaling Up Dense and Sequence in
  Industrial Recommenders}. In \bibinfo{booktitle}{\emph{Proceedings of the
  32nd ACM SIGKDD Conference on Knowledge Discovery and Data Mining, Volume
  2}}. \bibinfo{publisher}{ACM}, \bibinfo{pages}{7422--7431}.
\newblock
\urldef\tempurl%
\url{https://doi.org/10.1145/3770855.3818447}
\showDOI{\tempurl}


\bibitem[Kaplan et~al\mbox{.}(2020)]%
        {kaplan2020scaling}
\bibfield{author}{\bibinfo{person}{Jared Kaplan}, \bibinfo{person}{Sam
  McCandlish}, \bibinfo{person}{Tom Henighan}, \bibinfo{person}{Tom~B. Brown},
  \bibinfo{person}{Benjamin Chess}, \bibinfo{person}{Rewon Child},
  \bibinfo{person}{Scott Gray}, \bibinfo{person}{Alec Radford},
  \bibinfo{person}{Jeffrey Wu}, {and} \bibinfo{person}{Dario Amodei}.}
  \bibinfo{year}{2020}\natexlab{}.
\newblock \bibinfo{title}{Scaling Laws for Neural Language Models}.
\newblock
\newblock
\showeprint[arxiv]{2001.08361}~[cs.LG]
\urldef\tempurl%
\url{https://arxiv.org/abs/2001.08361}
\showURL{%
\tempurl}


\bibitem[Kennedy and O'Hagan(2000)]%
        {kennedy2000predicting}
\bibfield{author}{\bibinfo{person}{M.~C. Kennedy} {and} \bibinfo{person}{A.
  O'Hagan}.} \bibinfo{year}{2000}\natexlab{}.
\newblock \showarticletitle{Predicting the Output from a Complex Computer Code
  When Fast Approximations Are Available}.
\newblock \bibinfo{journal}{\emph{Biometrika}} \bibinfo{volume}{87},
  \bibinfo{number}{1} (\bibinfo{year}{2000}), \bibinfo{pages}{1--13}.
\newblock
\urldef\tempurl%
\url{https://doi.org/10.1093/biomet/87.1.1}
\showDOI{\tempurl}


\bibitem[Klein et~al\mbox{.}(2017)]%
        {klein2017fast}
\bibfield{author}{\bibinfo{person}{Aaron Klein}, \bibinfo{person}{Stefan
  Falkner}, \bibinfo{person}{Simon Bartels}, \bibinfo{person}{Philipp Hennig},
  {and} \bibinfo{person}{Frank Hutter}.} \bibinfo{year}{2017}\natexlab{}.
\newblock \showarticletitle{Fast {Bayesian} Optimization of Machine Learning
  Hyperparameters on Large Datasets}. In \bibinfo{booktitle}{\emph{Proceedings
  of the 20th International Conference on Artificial Intelligence and
  Statistics}} \emph{(\bibinfo{series}{Proceedings of Machine Learning
  Research}, Vol.~\bibinfo{volume}{54})}. \bibinfo{publisher}{PMLR},
  \bibinfo{pages}{528--536}.
\newblock
\urldef\tempurl%
\url{https://proceedings.mlr.press/v54/klein17a.html}
\showURL{%
\tempurl}


\bibitem[Letham et~al\mbox{.}(2020)]%
        {letham2020alebo}
\bibfield{author}{\bibinfo{person}{Benjamin Letham}, \bibinfo{person}{Roberto
  Calandra}, \bibinfo{person}{Akshara Rai}, {and} \bibinfo{person}{Eytan
  Bakshy}.} \bibinfo{year}{2020}\natexlab{}.
\newblock \showarticletitle{Re-Examining Linear Embeddings for High-Dimensional
  {Bayesian} Optimization}. In \bibinfo{booktitle}{\emph{Advances in Neural
  Information Processing Systems}}, Vol.~\bibinfo{volume}{33}.
  \bibinfo{publisher}{Curran Associates, Inc.}, \bibinfo{pages}{1546--1558}.
\newblock
\urldef\tempurl%
\url{https://arxiv.org/abs/2001.11659}
\showURL{%
\tempurl}


\bibitem[Li et~al\mbox{.}(2017)]%
        {li2017dropout}
\bibfield{author}{\bibinfo{person}{Cheng Li}, \bibinfo{person}{Sunil Gupta},
  \bibinfo{person}{Santu Rana}, \bibinfo{person}{Vu Nguyen},
  \bibinfo{person}{Svetha Venkatesh}, {and} \bibinfo{person}{Alistair
  Shilton}.} \bibinfo{year}{2017}\natexlab{}.
\newblock \showarticletitle{High dimensional {Bayesian} optimization using
  dropout}. In \bibinfo{booktitle}{\emph{Proceedings of the 26th International
  Joint Conference on Artificial Intelligence}}. \bibinfo{pages}{2096--2102}.
\newblock


\bibitem[Li et~al\mbox{.}(2025)]%
        {li2025misfitting}
\bibfield{author}{\bibinfo{person}{Margaret Li}, \bibinfo{person}{Sneha
  Kudugunta}, {and} \bibinfo{person}{Luke Zettlemoyer}.}
  \bibinfo{year}{2025}\natexlab{}.
\newblock \showarticletitle{{(Mis)Fitting: A Survey of Scaling Laws}}. In
  \bibinfo{booktitle}{\emph{The Thirteenth International Conference on Learning
  Representations}}.
\newblock
\showeprint[arxiv]{2502.18969}
\urldef\tempurl%
\url{https://openreview.net/forum?id=xI71dsS3o4}
\showURL{%
\tempurl}


\bibitem[Li et~al\mbox{.}(2026)]%
        {li2026spend}
\bibfield{author}{\bibinfo{person}{Sijie Li}, \bibinfo{person}{Shanda Li},
  \bibinfo{person}{Haowei Lin}, \bibinfo{person}{Weiwei Sun},
  \bibinfo{person}{Ameet Talwalkar}, {and} \bibinfo{person}{Yiming Yang}.}
  \bibinfo{year}{2026}\natexlab{}.
\newblock \bibinfo{title}{Spend Less, Fit Better: Budget-Efficient Scaling Law
  Fitting via Active Experiment Selection}.
\newblock
\newblock
\showeprint[arxiv]{2604.22753}~[cs.LG]
\urldef\tempurl%
\url{https://arxiv.org/abs/2604.22753}
\showURL{%
\tempurl}


\bibitem[Liu et~al\mbox{.}(2026)]%
        {liu2026template}
\bibfield{author}{\bibinfo{person}{Jiang Liu}, \bibinfo{person}{John~Martabano
  Landy}, \bibinfo{person}{Yao Xuan}, \bibinfo{person}{Djordje Gligorijevic},
  \bibinfo{person}{Swamy Muddu}, \bibinfo{person}{Nhat Le}, {et~al\mbox{.}}}
  \bibinfo{year}{2026}\natexlab{}.
\newblock \showarticletitle{Template-Driven Machine Learning Standardization
  for Large Model Fleets}. In \bibinfo{booktitle}{\emph{Proceedings of the 2026
  IEEE International Conference on Big Data}}. \bibinfo{publisher}{IEEE},
  \bibinfo{address}{Phoenix, AZ, USA}.
\newblock
\showeprint[arxiv]{2603.24963}
\newblock
\shownote{To appear}.


\bibitem[McCandlish et~al\mbox{.}(2018)]%
        {mccandlish2018largebatch}
\bibfield{author}{\bibinfo{person}{Sam McCandlish}, \bibinfo{person}{Jared
  Kaplan}, \bibinfo{person}{Dario Amodei}, {and} \bibinfo{person}{{OpenAI Dota
  Team}}.} \bibinfo{year}{2018}\natexlab{}.
\newblock \showarticletitle{An Empirical Model of Large-Batch Training}.
\newblock  (\bibinfo{year}{2018}).
\newblock
\urldef\tempurl%
\url{https://arxiv.org/abs/1812.06162}
\showURL{%
\tempurl}


\bibitem[Naumov et~al\mbox{.}(2019)]%
        {naumov2019dlrm}
\bibfield{author}{\bibinfo{person}{Maxim Naumov}, \bibinfo{person}{Dheevatsa
  Mudigere}, \bibinfo{person}{Hao-Jun~Michael Shi}, \bibinfo{person}{Jianyu
  Huang}, \bibinfo{person}{Narayanan Sundaraman}, \bibinfo{person}{Jongsoo
  Park}, \bibinfo{person}{Xiaodong Wang}, \bibinfo{person}{Udit Gupta},
  \bibinfo{person}{Carole-Jean Wu}, \bibinfo{person}{Alisson~G. Azzolini},
  \bibinfo{person}{Dmytro Dzhulgakov}, \bibinfo{person}{Andrey Mallevich},
  \bibinfo{person}{Ilia Cherniavskii}, \bibinfo{person}{Yinghai Lu},
  \bibinfo{person}{Raghuraman Krishnamoorthi}, \bibinfo{person}{Ansha Yu},
  \bibinfo{person}{Volodymyr Kondratenko}, \bibinfo{person}{Stephanie Pereira},
  \bibinfo{person}{Xianjie Chen}, \bibinfo{person}{Wenlin Chen},
  \bibinfo{person}{Vijay Rao}, \bibinfo{person}{Bill Jia},
  \bibinfo{person}{Liang Xiong}, {and} \bibinfo{person}{Misha Smelyanskiy}.}
  \bibinfo{year}{2019}\natexlab{}.
\newblock \bibinfo{title}{Deep Learning Recommendation Model for
  Personalization and Recommendation Systems}.
\newblock
\newblock
\urldef\tempurl%
\url{https://doi.org/10.48550/arXiv.1906.00091}
\showDOI{\tempurl}
\showeprint[arxiv]{1906.00091}~[cs.IR]


\bibitem[Olson et~al\mbox{.}(2025)]%
        {olson2025ax}
\bibfield{author}{\bibinfo{person}{Miles Olson}, \bibinfo{person}{Elizabeth
  Santorella}, \bibinfo{person}{Louis~C. Tiao}, \bibinfo{person}{Sait Cakmak},
  \bibinfo{person}{Mia Garrard}, \bibinfo{person}{Samuel Daulton},
  \bibinfo{person}{Zhiyuan~Jerry Lin}, \bibinfo{person}{Sebastian Ament},
  \bibinfo{person}{Bernard Beckerman}, \bibinfo{person}{Eric Onofrey},
  \bibinfo{person}{Paschal Igusti}, \bibinfo{person}{Cristian Lara},
  \bibinfo{person}{Benjamin Letham}, \bibinfo{person}{Cesar Cardoso},
  \bibinfo{person}{Shiyun~Sunny Shen}, \bibinfo{person}{Andy~Chenyuan Lin},
  \bibinfo{person}{Matthew Grange}, \bibinfo{person}{Elena Kashtelyan},
  \bibinfo{person}{David Eriksson}, \bibinfo{person}{Maximilian Balandat},
  {and} \bibinfo{person}{Eytan Bakshy}.} \bibinfo{year}{2025}\natexlab{}.
\newblock \showarticletitle{Ax: A Platform for Adaptive Experimentation}. In
  \bibinfo{booktitle}{\emph{Proceedings of the Fourth International Conference
  on Automated Machine Learning}} \emph{(\bibinfo{series}{Proceedings of
  Machine Learning Research}, Vol.~\bibinfo{volume}{293})}.
  \bibinfo{publisher}{PMLR}, \bibinfo{pages}{21/1--25}.
\newblock
\urldef\tempurl%
\url{https://proceedings.mlr.press/v293/olson25a.html}
\showURL{%
\tempurl}


\bibitem[Shen et~al\mbox{.}(2024)]%
        {shen2024optimizing}
\bibfield{author}{\bibinfo{person}{Tingjia Shen}, \bibinfo{person}{Hao Wang},
  \bibinfo{person}{Chuhan Wu}, \bibinfo{person}{Jin~Yao Chin},
  \bibinfo{person}{Wei Guo}, \bibinfo{person}{Yong Liu},
  \bibinfo{person}{Huifeng Guo}, \bibinfo{person}{Defu Lian},
  \bibinfo{person}{Ruiming Tang}, {and} \bibinfo{person}{Enhong Chen}.}
  \bibinfo{year}{2024}\natexlab{}.
\newblock \showarticletitle{Optimizing sequential recommendation models with
  scaling laws and approximate entropy}.
\newblock \bibinfo{journal}{\emph{arXiv preprint arXiv:2412.00430}}
  (\bibinfo{year}{2024}).
\newblock


\bibitem[Shen et~al\mbox{.}(2025)]%
        {shen2025plaw}
\bibfield{author}{\bibinfo{person}{Tingjia Shen}, \bibinfo{person}{Hao Wang},
  \bibinfo{person}{Chuhan Wu}, \bibinfo{person}{Jin~Yao Chin},
  \bibinfo{person}{Wei Guo}, \bibinfo{person}{Yong Liu},
  \bibinfo{person}{Huifeng Guo}, \bibinfo{person}{Defu Lian},
  \bibinfo{person}{Ruiming Tang}, {and} \bibinfo{person}{Enhong Chen}.}
  \bibinfo{year}{2025}\natexlab{}.
\newblock \showarticletitle{{P-Law: Predicting Quantitative Scaling Law with
  Entropy Guidance in Large Recommendation Models}}. In
  \bibinfo{booktitle}{\emph{Advances in Neural Information Processing
  Systems}}, Vol.~\bibinfo{volume}{38}. \bibinfo{publisher}{Curran Associates,
  Inc.}, \bibinfo{pages}{12911--12940}.
\newblock
\urldef\tempurl%
\url{https://doi.org/10.52202/085713-0436}
\showDOI{\tempurl}


\bibitem[Snoek et~al\mbox{.}(2012)]%
        {snoek2012practical}
\bibfield{author}{\bibinfo{person}{Jasper Snoek}, \bibinfo{person}{Hugo
  Larochelle}, {and} \bibinfo{person}{Ryan~P. Adams}.}
  \bibinfo{year}{2012}\natexlab{}.
\newblock \showarticletitle{Practical {Bayesian} Optimization of Machine
  Learning Algorithms}. In \bibinfo{booktitle}{\emph{Advances in Neural
  Information Processing Systems}}, Vol.~\bibinfo{volume}{25}.
  \bibinfo{publisher}{Curran Associates, Inc.}, \bibinfo{pages}{2951--2959}.
\newblock
\urldef\tempurl%
\url{https://proceedings.neurips.cc/paper/2012/hash/05311655a15b75fab86956663e1819cd-Abstract.html}
\showURL{%
\tempurl}


\bibitem[Swersky et~al\mbox{.}(2014)]%
        {swersky2014freezethaw}
\bibfield{author}{\bibinfo{person}{Kevin Swersky}, \bibinfo{person}{Jasper
  Snoek}, {and} \bibinfo{person}{Ryan~P. Adams}.}
  \bibinfo{year}{2014}\natexlab{}.
\newblock \showarticletitle{Freeze-Thaw Bayesian Optimization}.
\newblock \bibinfo{journal}{\emph{arXiv preprint arXiv:1406.3896}}
  (\bibinfo{year}{2014}).
\newblock


\bibitem[Wang et~al\mbox{.}(2024)]%
        {wang2024scalinglaws}
\bibfield{author}{\bibinfo{person}{Yunli Wang}, \bibinfo{person}{Zhen Zhang},
  \bibinfo{person}{Zixuan Yang}, \bibinfo{person}{Tianyu Xu},
  \bibinfo{person}{Zhiqiang Wang}, \bibinfo{person}{Yu Li},
  \bibinfo{person}{Rufan Zhou}, \bibinfo{person}{Zhiqiang Liu},
  \bibinfo{person}{Yanjie Zhu}, \bibinfo{person}{Jian Yang},
  \bibinfo{person}{Shiyang Wen}, {and} \bibinfo{person}{Peng Jiang}.}
  \bibinfo{year}{2024}\natexlab{}.
\newblock \bibinfo{title}{Scaling Laws for Online Advertisement Retrieval}.
\newblock
\newblock
\urldef\tempurl%
\url{https://doi.org/10.48550/arXiv.2411.13322}
\showDOI{\tempurl}
\showeprint[arxiv]{2411.13322}~[cs.IR]


\bibitem[Wang et~al\mbox{.}(2016)]%
        {wang2016rembo}
\bibfield{author}{\bibinfo{person}{Ziyu Wang}, \bibinfo{person}{Frank Hutter},
  \bibinfo{person}{Masrour Zoghi}, \bibinfo{person}{David Matheson}, {and}
  \bibinfo{person}{Nando de Freitas}.} \bibinfo{year}{2016}\natexlab{}.
\newblock \showarticletitle{Bayesian optimization in a billion dimensions via
  random embeddings}.
\newblock \bibinfo{journal}{\emph{Journal of Artificial Intelligence Research}}
  \bibinfo{volume}{55}, \bibinfo{number}{1} (\bibinfo{year}{2016}),
  \bibinfo{pages}{361--387}.
\newblock
\urldef\tempurl%
\url{https://arxiv.org/abs/1301.1942}
\showURL{%
\tempurl}


\bibitem[Wu et~al\mbox{.}(2020)]%
        {wu2020practical}
\bibfield{author}{\bibinfo{person}{Jian Wu}, \bibinfo{person}{Saul
  Toscano-Palmerin}, \bibinfo{person}{Peter~I. Frazier}, {and}
  \bibinfo{person}{Andrew~Gordon Wilson}.} \bibinfo{year}{2020}\natexlab{}.
\newblock \showarticletitle{Practical Multi-fidelity {Bayesian} Optimization
  for Hyperparameter Tuning}. In \bibinfo{booktitle}{\emph{Proceedings of the
  35th Uncertainty in Artificial Intelligence Conference}}
  \emph{(\bibinfo{series}{Proceedings of Machine Learning Research},
  Vol.~\bibinfo{volume}{115})}. \bibinfo{publisher}{PMLR},
  \bibinfo{pages}{788--798}.
\newblock
\urldef\tempurl%
\url{https://proceedings.mlr.press/v115/wu20a.html}
\showURL{%
\tempurl}


\bibitem[Xiong et~al\mbox{.}(2026)]%
        {xiong2026llatte}
\bibfield{author}{\bibinfo{person}{Lee Xiong}, \bibinfo{person}{Zhirong Chen},
  \bibinfo{person}{Rahul Mayuranath}, \bibinfo{person}{Shangran Qiu},
  \bibinfo{person}{Ahmet~Arda Ozdemir}, \bibinfo{person}{Lu Li},
  \bibinfo{person}{Yang Hu}, \bibinfo{person}{Dave Li},
  \bibinfo{person}{Jingtao Ren}, \bibinfo{person}{Howard Cheng},
  \bibinfo{person}{Fabian~Souto Herrera}, \bibinfo{person}{Ahmed Agiza},
  \bibinfo{person}{Allen Lin}, \bibinfo{person}{Baruch Epstein},
  \bibinfo{person}{Anuj Aggarwal}, \bibinfo{person}{Julia Ulziisaikhan},
  \bibinfo{person}{Chao Wang}, \bibinfo{person}{Dinesh Ramasamy},
  \bibinfo{person}{Parshva Doshi}, \bibinfo{person}{Sri Reddy}, {and}
  \bibinfo{person}{Arnold Overwijk}.} \bibinfo{year}{2026}\natexlab{}.
\newblock \showarticletitle{{LLaTTE: Scaling Laws for Multi-Stage Sequence
  Modeling in Large-Scale Ads Recommendation}}. In
  \bibinfo{booktitle}{\emph{Proceedings of the 32nd ACM SIGKDD Conference on
  Knowledge Discovery and Data Mining, Volume 2}}. \bibinfo{publisher}{ACM},
  \bibinfo{pages}{8301--8311}.
\newblock
\urldef\tempurl%
\url{https://doi.org/10.1145/3770855.3818442}
\showDOI{\tempurl}


\bibitem[Yang et~al\mbox{.}(2021)]%
        {yang2021tensorv}
\bibfield{author}{\bibinfo{person}{Greg Yang}, \bibinfo{person}{Edward~J. Hu},
  \bibinfo{person}{Igor Babuschkin}, \bibinfo{person}{Szymon Sidor},
  \bibinfo{person}{Xiaodong Liu}, \bibinfo{person}{David Farhi},
  \bibinfo{person}{Nick Ryder}, \bibinfo{person}{Jakub Pachocki},
  \bibinfo{person}{Weizhu Chen}, {and} \bibinfo{person}{Jianfeng Gao}.}
  \bibinfo{year}{2021}\natexlab{}.
\newblock \showarticletitle{Tensor Programs V: Tuning Large Neural Networks via
  Zero-Shot Hyperparameter Transfer}. In \bibinfo{booktitle}{\emph{Advances in
  Neural Information Processing Systems}}, Vol.~\bibinfo{volume}{34}.
\newblock
\showeprint[arxiv]{2203.03466}~[cs.LG]


\bibitem[Yang et~al\mbox{.}(2024)]%
        {yang2024tensorvi}
\bibfield{author}{\bibinfo{person}{Greg Yang}, \bibinfo{person}{Dingli Yu},
  \bibinfo{person}{Chen Zhu}, {and} \bibinfo{person}{Soufiane Hayou}.}
  \bibinfo{year}{2024}\natexlab{}.
\newblock \showarticletitle{Tensor Programs VI: Feature Learning in
  Infinite-Depth Neural Networks}. In \bibinfo{booktitle}{\emph{The Twelfth
  International Conference on Learning Representations (ICLR)}}.
\newblock
\showeprint[arxiv]{2310.02244}~[cs.LG]


\bibitem[Zeng et~al\mbox{.}(2025)]%
        {zeng2025interformer}
\bibfield{author}{\bibinfo{person}{Zhichen Zeng}, \bibinfo{person}{Xiaolong
  Liu}, \bibinfo{person}{Mengyue Hang}, \bibinfo{person}{Xiaoyi Liu},
  \bibinfo{person}{Qinghai Zhou}, \bibinfo{person}{Chaofei Yang},
  \bibinfo{person}{Yiqun Liu}, \bibinfo{person}{Yichen Ruan},
  \bibinfo{person}{Laming Chen}, \bibinfo{person}{Yuxin Chen},
  \bibinfo{person}{Yujia Hao}, \bibinfo{person}{Jiaqi Xu},
  \bibinfo{person}{Jade Nie}, \bibinfo{person}{Xi Liu}, \bibinfo{person}{Buyun
  Zhang}, \bibinfo{person}{Wei Wen}, \bibinfo{person}{Siyang Yuan},
  \bibinfo{person}{Hang Yin}, \bibinfo{person}{Xin Zhang}, \bibinfo{person}{Kai
  Wang}, \bibinfo{person}{Wen-Yen Chen}, \bibinfo{person}{Yiping Han},
  \bibinfo{person}{Huayu Li}, \bibinfo{person}{Chunzhi Yang},
  \bibinfo{person}{Bo Long}, \bibinfo{person}{Philip~S. Yu},
  \bibinfo{person}{Hanghang Tong}, {and} \bibinfo{person}{Jiyan Yang}.}
  \bibinfo{year}{2025}\natexlab{}.
\newblock \showarticletitle{{InterFormer}: Effective Heterogeneous Interaction
  Learning for Click-Through Rate Prediction}. In
  \bibinfo{booktitle}{\emph{Proceedings of the 34th ACM International
  Conference on Information and Knowledge Management}}.
  \bibinfo{publisher}{ACM}, \bibinfo{pages}{6225--6233}.
\newblock
\urldef\tempurl%
\url{https://doi.org/10.1145/3746252.3761527}
\showDOI{\tempurl}


\bibitem[Zhai et~al\mbox{.}(2024)]%
        {zhai2024actions}
\bibfield{author}{\bibinfo{person}{Jiaqi Zhai}, \bibinfo{person}{Lucy Liao},
  \bibinfo{person}{Xing Liu}, \bibinfo{person}{Yueming Wang},
  \bibinfo{person}{Rui Li}, \bibinfo{person}{Xuan Cao}, \bibinfo{person}{Leon
  Gao}, \bibinfo{person}{Zhaojie Gong}, \bibinfo{person}{Fangda Gu},
  \bibinfo{person}{Jiayuan He}, \bibinfo{person}{Yinghai Lu}, {and}
  \bibinfo{person}{Yu Shi}.} \bibinfo{year}{2024}\natexlab{}.
\newblock \showarticletitle{Actions Speak Louder than Words: Trillion-Parameter
  Sequential Transducers for Generative Recommendations}. In
  \bibinfo{booktitle}{\emph{Proceedings of the 41st International Conference on
  Machine Learning}} \emph{(\bibinfo{series}{Proceedings of Machine Learning
  Research}, Vol.~\bibinfo{volume}{235})}. \bibinfo{publisher}{PMLR},
  \bibinfo{pages}{58484--58509}.
\newblock
\urldef\tempurl%
\url{https://proceedings.mlr.press/v235/zhai24a.html}
\showURL{%
\tempurl}


\bibitem[Zhang et~al\mbox{.}(2024)]%
        {zhang2024wukong}
\bibfield{author}{\bibinfo{person}{Buyun Zhang}, \bibinfo{person}{Liang Luo},
  \bibinfo{person}{Yuxin Chen}, \bibinfo{person}{Jade Nie}, \bibinfo{person}{Xi
  Liu}, \bibinfo{person}{Shen Li}, \bibinfo{person}{Yanli Zhao},
  \bibinfo{person}{Yuchen Hao}, \bibinfo{person}{Yantao Yao},
  \bibinfo{person}{Ellie~Dingqiao Wen}, \bibinfo{person}{Jongsoo Park},
  \bibinfo{person}{Maxim Naumov}, {and} \bibinfo{person}{Wenlin Chen}.}
  \bibinfo{year}{2024}\natexlab{}.
\newblock \showarticletitle{{Wukong: Towards a Scaling Law for Large-Scale
  Recommendation}}. In \bibinfo{booktitle}{\emph{Proceedings of the 41st
  International Conference on Machine Learning}}
  \emph{(\bibinfo{series}{Proceedings of Machine Learning Research},
  Vol.~\bibinfo{volume}{235})}. \bibinfo{publisher}{PMLR},
  \bibinfo{pages}{59421--59434}.
\newblock
\urldef\tempurl%
\url{https://proceedings.mlr.press/v235/zhang24ao.html}
\showURL{%
\tempurl}


\bibitem[Zhang et~al\mbox{.}(2022)]%
        {zhang2022dhen}
\bibfield{author}{\bibinfo{person}{Buyun Zhang}, \bibinfo{person}{Liang Luo},
  \bibinfo{person}{Xi Liu}, \bibinfo{person}{Jay Li}, \bibinfo{person}{Zeliang
  Chen}, \bibinfo{person}{Weilin Zhang}, \bibinfo{person}{Xiaohan Wei},
  \bibinfo{person}{Yuchen Hao}, \bibinfo{person}{Michael Tsang},
  \bibinfo{person}{Wenjun Wang}, \bibinfo{person}{Yang Liu},
  \bibinfo{person}{Huayu Li}, \bibinfo{person}{Yasmine Badr},
  \bibinfo{person}{Jongsoo Park}, \bibinfo{person}{Jiyan Yang},
  \bibinfo{person}{Dheevatsa Mudigere}, {and} \bibinfo{person}{Ellie Wen}.}
  \bibinfo{year}{2022}\natexlab{}.
\newblock \bibinfo{title}{{DHEN}: A Deep and Hierarchical Ensemble Network for
  Large-Scale Click-Through Rate Prediction}.
\newblock
\newblock
\urldef\tempurl%
\url{https://doi.org/10.48550/arXiv.2203.11014}
\showDOI{\tempurl}
\showeprint[arxiv]{2203.11014}~[cs.IR]


\bibitem[Zitzler et~al\mbox{.}(2007)]%
        {zitzler2007thehypervolume}
\bibfield{author}{\bibinfo{person}{Eckart Zitzler}, \bibinfo{person}{Dimo
  Brockhoff}, {and} \bibinfo{person}{Lothar Thiele}.}
  \bibinfo{year}{2007}\natexlab{}.
\newblock \showarticletitle{The Hypervolume Indicator Revisited: On the Design
  of Pareto-compliant Indicators Via Weighted Integration}. In
  \bibinfo{booktitle}{\emph{Evolutionary Multi-Criterion Optimization}}.
  \bibinfo{publisher}{Springer Berlin Heidelberg}, \bibinfo{address}{Berlin,
  Heidelberg}, \bibinfo{pages}{862--876}.
\newblock
\showISBNx{978-3-540-70928-2}


\bibitem[Zitzler and Thiele(1999)]%
        {zitzler1999multiobjective}
\bibfield{author}{\bibinfo{person}{Eckart Zitzler} {and}
  \bibinfo{person}{Lothar Thiele}.} \bibinfo{year}{1999}\natexlab{}.
\newblock \showarticletitle{Multiobjective evolutionary algorithms: a
  comparative case study and the strength Pareto approach}. In
  \bibinfo{booktitle}{\emph{IEEE Transactions on Evolutionary Computation}},
  Vol.~\bibinfo{volume}{3}. \bibinfo{publisher}{IEEE},
  \bibinfo{pages}{257--271}.
\newblock


\end{thebibliography}

\appendix 
\section{Training Parameter Pre-Optimization} 
\label{app:training_params} 
Batch size and learning rate interact with model capacity, so an optimal configuration may differ between small and large models. In the language-model setting this dependence is well characterized as a function of a single scalar notion of size---optimal batch size tracks a scalar summary of the training problem~\citep{mccandlish2018largebatch}, optimal hyperparameters follow power laws in model size and compute~\citep{kaplan2020scaling}, and $\mu$P permits zero-shot transfer across width~\citep{yang2021tensorv}---but extending transfer beyond one axis is not automatic: the analogous guarantee along depth required a separate theoretical treatment~\citep{yang2024tensorvi}. 

\begin{figure}[htbp]
    \centering
    \includegraphics[width=\columnwidth, keepaspectratio]{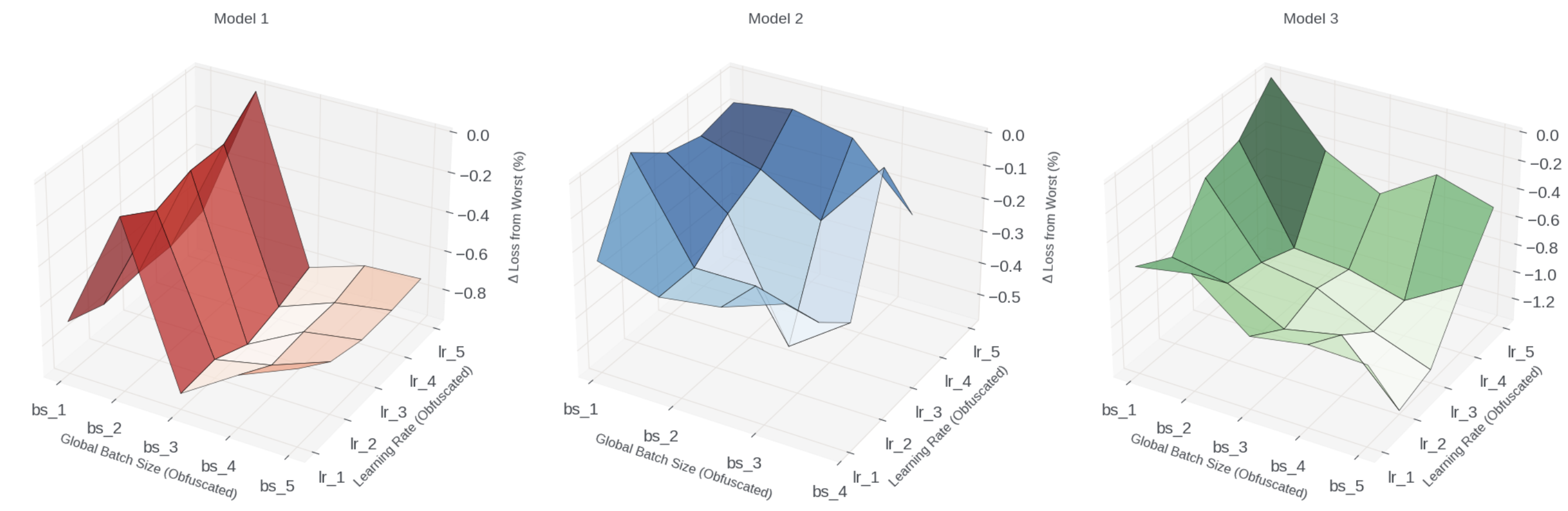}
    \caption{Tuning training configuration yields consistent and significant performance upside up to 1.2\% across models of varied data distributions and optimization events, with loss landscapes following a smooth, differentiable shape in the grid. Absolute batch-size and learning-rate values are omitted for confidentiality.}
    \label{fig:training_config_sizing}
\Description{Three surface plots show loss changes across batch-size and learning-rate grids for three models. Each surface contains a lower-loss region, with the location and depth of the minimum varying across models. Axis values are anonymized.}
\end{figure}

A scaling layer distributes capacity across $k$ coordinates which the engine moves independently. Thus there is no single scalar along which a transfer rule could be expressed, and the question reopens in a form specific to our setting: does one training-parameter scaling law suffice per template, or must it be refit for each direction in $Z_\phi$? A grid search over learning rate and batch size across three diverse ranking models yielded NE gains up to $1.2\%$ over each model's default configuration, with loss surfaces smooth and near-unimodal over the grid (Figure~\ref{fig:training_config_sizing}).

Because these surfaces are smooth and near-unimodal, a fitted schedule can stand in for a per-candidate sweep. Axia therefore adopts a two-stage approach: fit the training-parameter scaling law from the sampling trial data already collected for cost profiling, then apply the learned schedule during the surrogate-driven search rather than sweeping training parameters at each candidate configuration. Because Power-Law Entropy Search~\citep{chen2026efficiently} extracts these relationships from existing trial data, this adds no full training runs to the campaign budget. One caveat on interpretation: in recommendation models the effective data volume is not a token count but reflects the diversity of user--item interactions and sparse feature cardinality, so fitted exponents are not directly comparable to those from language-modeling studies.
 
\section{Additional Experimental Details}
\label{app:experiment_details}

\subsection{Measurements and Comparison Protocol}
All configurations are compared under the same weighted NE metric, training data, and budget, with a cost ceiling of 160 MFLOPs per example. Quality is relative change from a common reference (absolute values are confidential). Inference FLOPs are profiler-measured, and reported frontiers use these measured costs with observed quality rather than surrogate predictions. Hypervolume uses a common reference point of 160 MFLOPs and approximately $+0.5369\%$ relative NE, and is reported in $[\mathrm{NE}\,\Delta\%\times\mathrm{MFLOPs}]$.

The 3D scaling layers search depth $\ell\in\{1,\ldots,5\}$, embedding dimension $d\in\{16,32,\ldots,192\}$, and width multiplier $w$ from 0.02 to 2 in steps of 0.02. Profiling-based methods share the 64 cost-profile configurations, but use separate sets of 16 initialization training observations. Ax without cost profiling receives no preliminary cost observations and does not screen its initialization for cost feasibility. Each non-grid comparison therefore evaluates a complete search procedure at 20 completed training runs.

\subsection{Baseline Configurations}
\paragraph{Random (Sobol) and Grids.}
Random search draws 20 distinct configurations from a scrambled Sobol sequence mapped to the finite coordinate sets with deduplication and top-up; this mapping is separate from Ax's native Sobol encoding. The $2\times2\times5$ grid uses $\ell\in\{1,5\}$, $d\in\{16,192\}$, and $w\in\{0.02,0.5,1,1.5,2\}$. The $3\times3\times3$ grid uses $\ell\in\{1,3,5\}$, $d\in\{16,104,192\}$, and $w\in\{0.02,1,2\}$. The grids share 12 physical configurations and their bounds match the adaptive search, though the second grid's embedding midpoint is off its discrete lattice. Neither grid is screened by a learned cost model before training.

\paragraph{BO (Ax).}
Ax is run at its shipped defaults, so the comparison reflects the platform as deployed: stock Ax already supplies informative priors and constraint-aware transformations. We use the default surrogate and acquisition dispatch of the evaluated internal build~\citep{olson2025ax} with this study's search space and 16+4 budget, initialized from 16 unfiltered native Sobol proposals; the four adaptive candidates are generated once those observations arrive. 

For a numeric space the default is a separate single-task GP per output with learned constant mean and ARD RBF kernel, point fitted under the platform's default lengthscale and noise priors; outputs pass through winsorization, a constraint-centred bilogarithmic transform for FLOPs, and standardization, with qLogNEI selected given a feasible history.

\paragraph{BO (Ax + Cost Profiling).}
This arm adds the shared profiling observations and cost-feasible initialization, rejecting Sobol candidates whose cost-surrogate mean exceeds the cap (up to 100 attempts each), then searches with the platform's default transfer-enabled modular strategy rather than Axia's kernel and transform settings. Its source/target mechanism models the profiling-to-training relationship separately per output. One implementation defect applies to this arm alone: the frozen source/target search-space merge reconstructs numeric choice parameters without preserving the target width's logarithmic flag. Axia corrects this, so the gap between the two plausibly overstates the contribution of Axia's remaining changes.

\paragraph{Open-Loop Surrogate.}
The predecessor fits point-prediction response models --- tree ensembles for the task-specific NE responses, linear regression for FLOPs --- on the 16 initialization trials, fits a scaling curve to the predicted complexity--quality frontier, and allocates four follow-up candidates from those point predictions with no uncertainty-based acquisition. Candidates are filtered to a neighborhood of the predicted frontier, then selected near the cost target and across the predicted-cost range.

\paragraph{Axia.}
The main figures use the input transform of Equation~\eqref{eq:input_transform} with qLogNEI. The GP directly fits observed weighted NE instead of reconstructing it from separately fitted heads. FLOPs alone receive a logarithmic output transform following the inherited cost winsorization, and both outputs are standardized. Separate single-task GPs use learned constant means with independent parameters for scaled ARD RBF and ARD linear components; the custom configuration adds no priors on lengthscales, outputscale, or linear variances, retaining the standard noise prior. Hyperparameters are point fitted. The production fit contains 16 weighted-NE rows and 78 deduplicated cost rows from profiling and training. The paired qLogNEHVI arm uses the same fitted surrogate and initialization, changing only the acquisition; its acquisition reference is fixed before proposal generation and differs from the common reference used for reporting hypervolume. Comparisons within each scheduling panel hold surrogate and acquisition fixed.

\paragraph{Acquisition implementation.}
\label{app:acquisition_details}
Constrained EI is implemented as qLogNEI and EHVI as qLogNEHVI, the prefixes denoting batch selection ($q$), a stable logarithmic formulation (Log), and noisy observations (N)~\citep{ament2023unexpected,daulton2021parallel}. Both are evaluated by posterior Monte Carlo sampling that integrates latent-outcome uncertainty at observed configurations, with feasibility folded into the sampled utility. The distinction between selecting one operating point and mapping scale-up tradeoffs does not depend on these choices.

\paragraph{Version boundary.}
The baseline defaults above refer to the evaluated Ax release 1.3.1, rather than the latest release. The experiment inventory retains kernel identities, serialized generator settings, model states, and per-attempt training-package pins.

\paragraph{Surrogate diagnostics.}
Aggregate fit statistics are an imperfect proxy for selection quality: RMSE and $R^2$ weight regions unlikely to be selected, and rank correlations weight every position equally, penalising a displacement near the top of the ranking no more than one at the bottom. A good aggregate score therefore need not identify a good batch under the FLOPs cap, and surrogate comparisons should report selection outcomes — best feasible NE and attained hypervolume for the selected batch, counting infeasible selections — alongside fit.

\section{Extensions}
\label{app:extensions}
The controlled studies in this work use completed training observations, independently modeled outputs, and an overall inference-cost constraint. The following extensions broaden how the system can reuse observations and allocate training and serving resources.

\subsection{Sharing Information Between NE and FLOPs}
For the output order $(f_{\NE},f_{\log\COST})$, Equation~\eqref{eq:hierarchical} gives covariance
\begin{equation}
\label{eq:joint_cov}
\mathbf{K}(\mathbf{u},\mathbf{u}')=\begin{bmatrix}
\beta^2 k_F(\mathbf{u},\mathbf{u}')+k_N(\mathbf{u},\mathbf{u}') & \beta k_F(\mathbf{u},\mathbf{u}')\\
\beta k_F(\mathbf{u},\mathbf{u}') & k_F(\mathbf{u},\mathbf{u}')
\end{bmatrix}.
\end{equation}
Affine response standardization is suppressed in this notation. The shared component lets profiling observations inform a quality trend while retaining architecture-dependent residual variation -- an autoregressive multi-output construction related to multi-fidelity modeling~\citep{kennedy2000predicting}, except that the outputs here are different measured quantities rather than two fidelities of one response. Setting $\beta=0$ recovers the independent models used in the reported searches; offline comparisons of coupled forms, including shared-output kernels, found no consistent advantage in this setting. Coupling remains a direction for learning when profiling can improve quality-based selection.

\subsection{Transfer Across Experiments}
\label{app:transfer}
Historical observations can inform a new campaign when experiments share a parameterization. We investigated pairwise source--target transfer using task-specific means and a learned task covariance over an RBF-plus-linear kernel, which shares information across experiments separately from coupling NE and FLOPs within one. Across offline studies with 2--16 target observations, suitable sources improved held-out prediction while less suitable sources degraded it --- motivating learning to select complementary sources, so that completed campaigns can reduce the training runs a new search requires.

\subsection{Training Progression}
\label{app:progression}
\begin{figure}[t]
\centering
\begingroup

\pgfplotsset{keep/.style 2 args={x filter/.append code={\edef\actual{\thisrow{#1}}\def\expected{#2}\ifx\actual\expected\else\def\pgfmathresult{nan}\fi}}}
\definecolor{halfcolor}{HTML}{9C7160}
\definecolor{finalcolor}{HTML}{315D83}

\begin{tikzpicture}
\begin{axis}[width=0.80\columnwidth,height=0.494\columnwidth,scale only axis,font=\small,
axis x line*=bottom,axis y line*=left,axis line style={black!50,line width=.35pt},
tick align=outside,tick style={black!50,line width=.35pt},tick label style={text=black!75},
ymajorgrids=true,major grid style={black!10,line width=.25pt},
xmin=15,xmax=180,ymin=-.55,ymax=.76,xtick={20,40,60,80,100,120,140,160,180},minor x tick num=1,
ytick={-.4,-.2,0,.2,.4,.6},scaled ticks=false,
yticklabel style={/pgf/number format/fixed,/pgf/number format/precision=1},
xlabel={Inference MFLOPs / example},ylabel={NE $\Delta\%$},
legend style={at={(.5,1.03)},anchor=south,draw=none,font=\small,cells={anchor=west},column sep=12pt},legend columns=2,
clip=true,unbounded coords=discard]
\draw[black!40,densely dotted,line width=.4pt] (axis cs:160,-.55)--(axis cs:160,.76);
\node[anchor=north east,text=black!55,font=\scriptsize] at (axis cs:159,.75) {160M cap};
\addplot[halfcolor,line width=.7pt,keep={snapshot}{half}] table[col sep=comma,x=mflops,y=ne_delta_percent] {figures/training-progression/frontier-paths.csv};
\addlegendentry{$\approx$50\% of examples}
\addplot[only marks,forget plot,halfcolor,mark=*,mark size=1.05pt,opacity=0.32,mark options={draw=white,line width=.2pt},keep={snapshot}{half},keep={efficient}{0}] table[col sep=comma,x=mflops,y=ne_delta_percent] {figures/training-progression/results-long.csv};
\addplot[only marks,forget plot,halfcolor,mark=*,mark size=1.05pt,opacity=1,mark options={draw=white,line width=.2pt},keep={snapshot}{half},keep={efficient}{1}] table[col sep=comma,x=mflops,y=ne_delta_percent] {figures/training-progression/results-long.csv};
\addplot[finalcolor,line width=.7pt,keep={snapshot}{final}] table[col sep=comma,x=mflops,y=ne_delta_percent] {figures/training-progression/frontier-paths.csv};
\addlegendentry{100\% of examples}
\addplot[only marks,forget plot,finalcolor,mark=*,mark size=1.05pt,opacity=0.32,mark options={draw=white,line width=.2pt},keep={snapshot}{final},keep={efficient}{0}] table[col sep=comma,x=mflops,y=ne_delta_percent] {figures/training-progression/results-long.csv};
\addplot[only marks,forget plot,finalcolor,mark=*,mark size=1.05pt,opacity=1,mark options={draw=white,line width=.2pt},keep={snapshot}{final},keep={efficient}{1}] table[col sep=comma,x=mflops,y=ne_delta_percent] {figures/training-progression/results-long.csv};
\end{axis}
\end{tikzpicture}

\endgroup
\caption{
Observed frontiers at $\sim$50\% and 100\% of processed training examples for the same 16 initialization and four adaptive architectures, the latter selected retrospectively from completed observations. Halfway records are the last aligned observation at or before the midpoint, not interpolated. Quality is cumulative NE relative to the reference; inference cost is as measured. Muted points are dominated within their snapshot.}
\label{fig:progression-production20}
\Description{NE versus inference FLOPs for the same twenty architectures at halfway and completed training. Blue final observations lie below their brown halfway counterparts. Staircases show changing frontier membership; muted points are dominated within each snapshot.}
\end{figure}
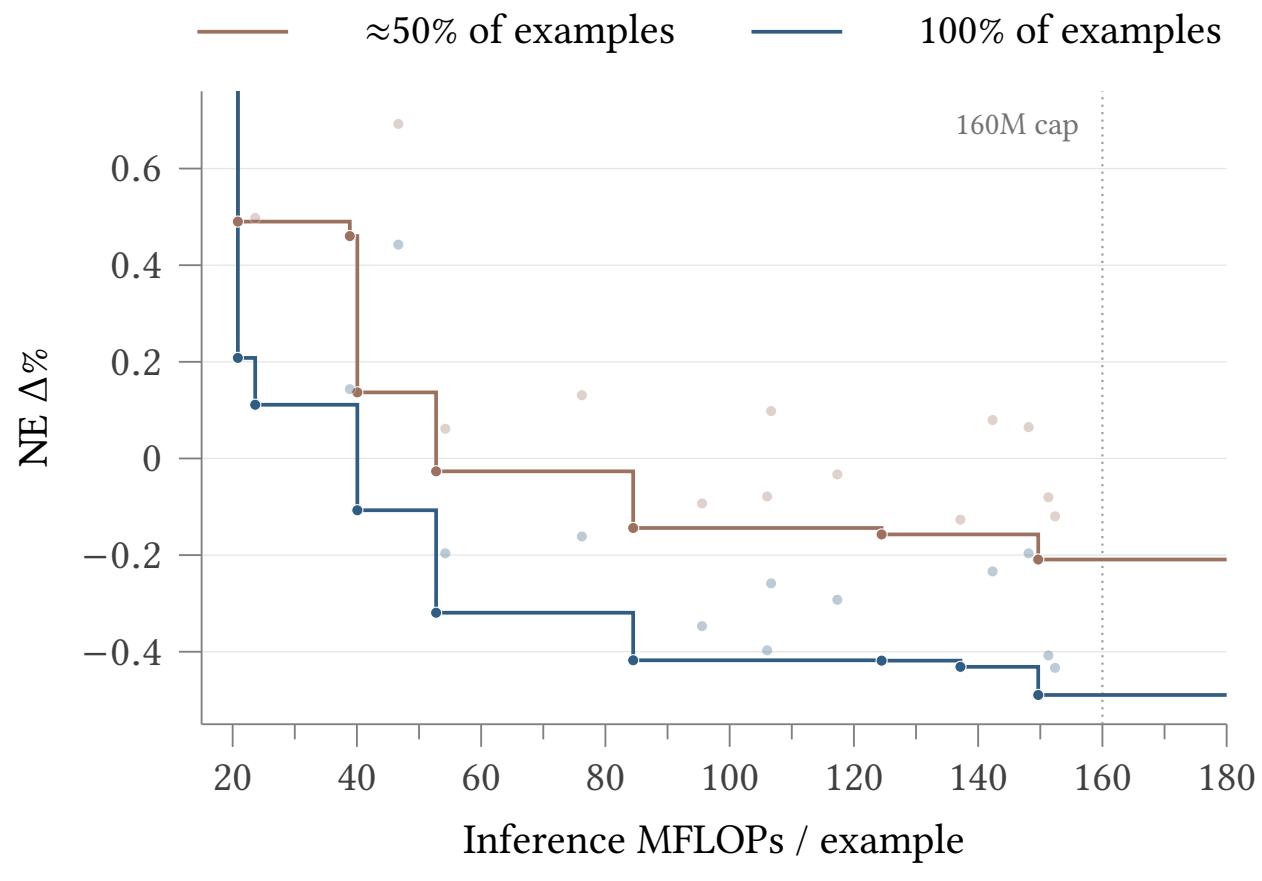

Progression-aware modeling is implemented and productionized: training progress enters the surrogate alongside architecture, so intermediate observations inform endpoint predictions and compute can be reallocated from unpromising configurations before they complete~\citep{domhan2015speeding,swersky2014freezethaw,falkner2018bohb}. Figure~\ref{fig:progression-production20} shows the frontier for the same 20 production 3D architectures at roughly half and full training: all 20 improve and the best feasible architecture is unchanged, but frontier membership shifts---early observations are informative, yet endpoint prediction is needed to track which architectures stay competitive.

\subsection{Candidate Allocation Under Different Acquisition Objectives}
\label{app:acquisition_comparison}
\begin{figure}[t]
\centering
\begingroup
\pgfplotsset{keep/.style 2 args={x filter/.append code={\edef\actual{\thisrow{#1}}\def\expected{#2}\ifx\actual\expected\else\def\pgfmathresult{nan}\fi}}}
\definecolor{init}{HTML}{9A9FA3}
\definecolor{ei}{HTML}{1267A2}
\definecolor{ehvi}{HTML}{258F83}
\begin{tikzpicture}
\begin{axis}[width=7.1cm,height=4.388cm,scale only axis,font=\small,axis x line*=bottom,axis y line*=left,axis line style={black!50,line width=.35pt},tick style={black!50,line width=.35pt},tick align=outside,major tick length=2.5pt,minor tick length=1.3pt,scaled ticks=false,tick label style={text=black!75},ymajorgrids=true,major grid style={black!10,line width=.25pt},xmin=0,xmax=180.0,ymin=-0.5494865959569895,ymax=0.5126071847724256,xtick={0,40,80,120,160},minor x tick num=1,ytick distance=.2,xlabel={Inference MFLOPs / example},ylabel={NE $\Delta\%$},unbounded coords=jump,legend style={at={(.5,-.22)},anchor=north,draw=none,font=\scriptsize,cells={anchor=west},column sep=5pt},legend columns=3]
\draw[black!40,densely dotted,line width=.5pt] (axis cs:160.0,-0.5494865959569895) -- (axis cs:160.0,0.5126071847724256);
\node[anchor=north east,font=\scriptsize,text=black!60] at (axis cs:160.0,0.5126071847724256) {160M cap};
\addplot[init,line width=.7pt,forget plot,keep={group}{Initialization},keep={segment}{0}] table[col sep=comma,x=mflops,y=ne_delta_percent] {figures/acquisition-comparison/paths-long.csv};
\addplot[ehvi,line width=.7pt,forget plot,keep={group}{EHVI},keep={segment}{0}] table[col sep=comma,x=mflops,y=ne_delta_percent] {figures/acquisition-comparison/paths-long.csv};
\addplot[ehvi,line width=.7pt,forget plot,keep={group}{EHVI},keep={segment}{1}] table[col sep=comma,x=mflops,y=ne_delta_percent] {figures/acquisition-comparison/paths-long.csv};
\addplot[ehvi,line width=.7pt,forget plot,keep={group}{EHVI},keep={segment}{2}] table[col sep=comma,x=mflops,y=ne_delta_percent] {figures/acquisition-comparison/paths-long.csv};
\addplot[ei,line width=.7pt,forget plot,keep={group}{EI},keep={segment}{0}] table[col sep=comma,x=mflops,y=ne_delta_percent] {figures/acquisition-comparison/paths-long.csv};
\addplot[ei,line width=.7pt,forget plot,keep={group}{EI},keep={segment}{1}] table[col sep=comma,x=mflops,y=ne_delta_percent] {figures/acquisition-comparison/paths-long.csv};
\addplot[only marks,init,mark=*,mark size=0.85pt,mark options={draw=white,line width=.2pt},keep={group}{Initialization}] table[col sep=comma,x=mflops,y=ne_delta_percent] {figures/acquisition-comparison/observations-long.csv};
\addlegendentry{Shared initialization}
\addplot[only marks,ei,mark=*,mark size=1.55pt,mark options={draw=white,line width=.2pt},keep={group}{EI}] table[col sep=comma,x=mflops,y=ne_delta_percent] {figures/acquisition-comparison/observations-long.csv};
\addlegendentry{Constrained EI}
\addplot[only marks,ehvi,mark=square*,mark size=1.4pt,mark options={draw=white,line width=.2pt},keep={group}{EHVI}] table[col sep=comma,x=mflops,y=ne_delta_percent] {figures/acquisition-comparison/observations-long.csv};
\addlegendentry{EHVI}
\end{axis}
\end{tikzpicture}
\endgroup
\caption{Candidate allocation with a shared initialization and surrogate. Gray points and steps show the 16 initialization observations and their frontier. Blue circles and green squares show the four configurations selected by constrained EI and EHVI, respectively; colored frontier segments show improvements over the initialization. NE is relative to a shared baseline; lower is better.}
\label{fig:acquisition-comparison}
\ifdefined\Description
\Description{A single NE versus inference-FLOPs panel. Constrained EI selects four architectures near the cost cap and improves the lowest NE. EHVI selects architectures across lower costs and expands more of the observed frontier.}
\fi
\end{figure}
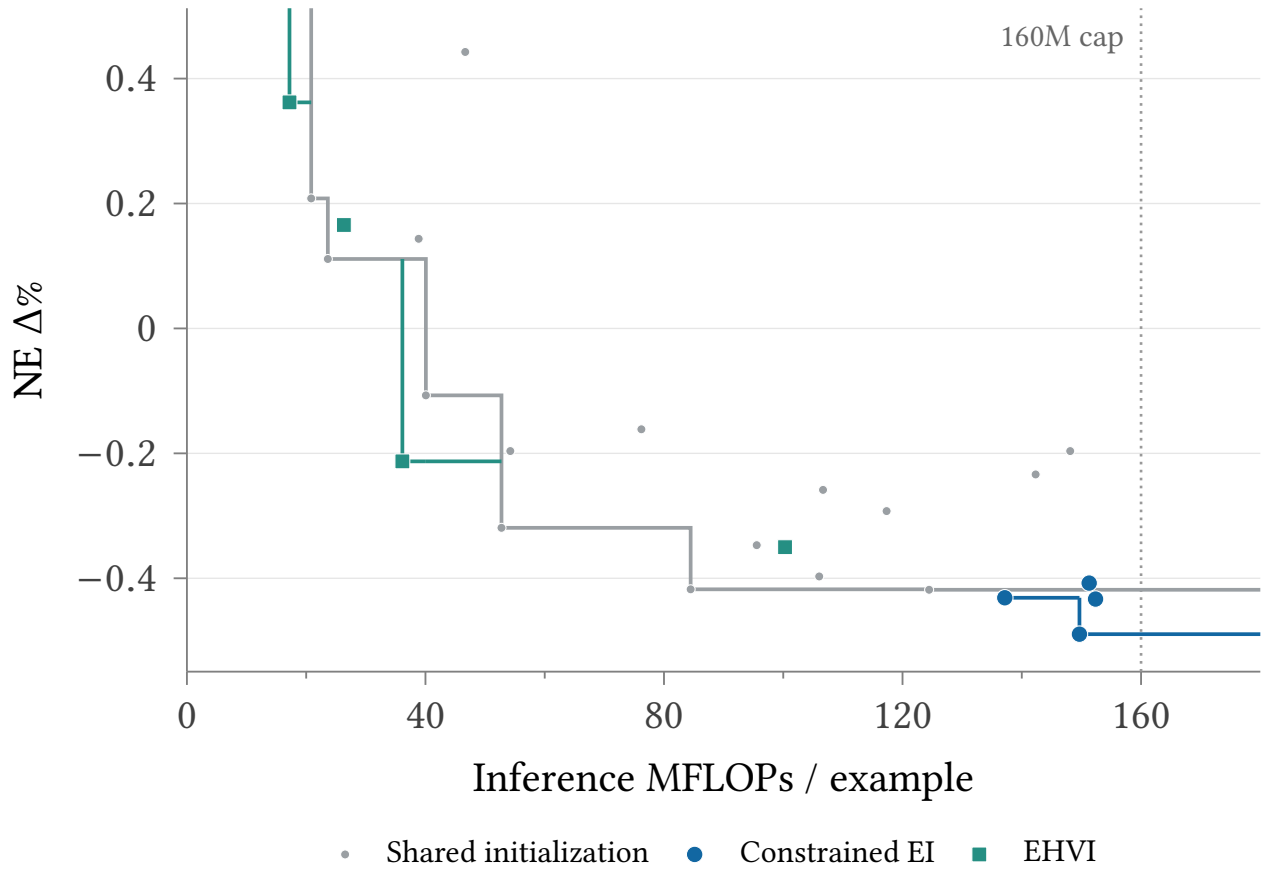

We compare constrained EI and EHVI from the same fitted surrogate and sixteen initialization trials for the production parameterization. Each acquisition selects four additional architectures. Constrained EI allocates its batch near the serving allowance and improves best feasible relative NE by 0.071 percentage points. EHVI explores lower-cost architectures and adds 3.6 times as much hypervolume, while retaining the initialization's best feasible NE (Figure~\ref{fig:acquisition-comparison}). The observed allocation patterns illustrate the different goals of optimizing quality within an allowance and exploring tradeoffs across the complexity--quality frontier.

\subsection{Predicting the Production Frontier}
\label{app:frontier_prediction}
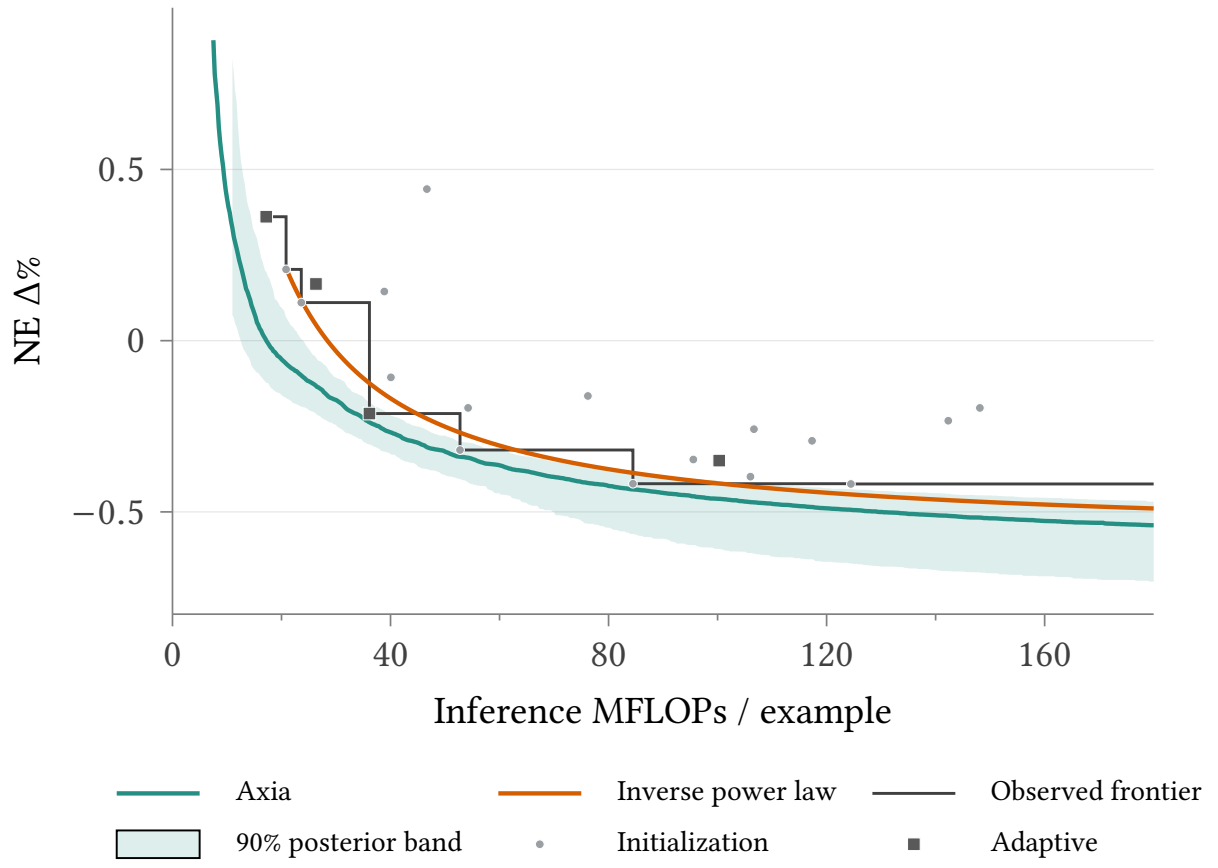
\begin{figure}[t]
\centering
\resizebox{\columnwidth}{!}{\begingroup
\pgfplotsset{keep/.style 2 args={x filter/.append code={\edef\actual{\thisrow{#1}}\def\expected{#2}\ifx\actual\expected\else\def\pgfmathresult{nan}\fi}}}
\definecolor{gp}{HTML}{258F83}
\definecolor{power}{HTML}{D55E00}
\definecolor{init}{HTML}{9A9FA3}
\begin{tikzpicture}
\begin{axis}[width=7.1cm,height=4.388cm,scale only axis,font=\small,axis x line*=bottom,axis y line*=left,axis line style={black!50,line width=.35pt},tick style={black!50,line width=.35pt},tick align=outside,major tick length=2.5pt,minor tick length=1.3pt,scaled ticks=false,tick label style={text=black!75},ymajorgrids=true,major grid style={black!10,line width=.25pt},xmin=0,xmax=180,ymin=-0.7986068649495826,ymax=0.9721574100032848,xtick={0,40,80,120,160},minor x tick num=1,xlabel={Inference MFLOPs / example},ylabel={NE $\Delta\%$},unbounded coords=jump,legend style={at={(.5,-.24)},anchor=north,draw=none,font=\scriptsize,cells={anchor=west},column sep=5pt,row sep=1pt},legend columns=3]
\addplot[name path=lower,draw=none,forget plot,restrict x to domain=11.0:180,keep={series}{gp_q05}] table[col sep=comma,x=mflops,y=ne_delta_percent] {figures/frontier-prediction/curves-long.csv};
\addplot[name path=upper,draw=none,forget plot,restrict x to domain=11.0:180,keep={series}{gp_q95}] table[col sep=comma,x=mflops,y=ne_delta_percent] {figures/frontier-prediction/curves-long.csv};
\addplot[gp,fill opacity=.15,draw=none,forget plot] fill between[of=lower and upper];
\addplot[black!75,line width=.6pt,forget plot] table[col sep=comma,x=mflops,y=ne_delta_percent] {figures/frontier-prediction/empirical-frontier.csv};
\addplot[gp,line width=.9pt,keep={series}{gp_median}] table[col sep=comma,x=mflops,y=ne_delta_percent] {figures/frontier-prediction/curves-long.csv};
\addlegendentry{Axia}
\addplot[power,line width=.9pt,keep={series}{inverse_power}] table[col sep=comma,x=mflops,y=ne_delta_percent] {figures/frontier-prediction/curves-long.csv};
\addlegendentry{Inverse power law}
\addlegendimage{black!75,line width=.6pt}
\addlegendentry{Observed frontier}
\addlegendimage{area legend,fill=gp,fill opacity=.15,draw=none}
\addlegendentry{90\% posterior band}
\addplot[only marks,init,mark=*,mark size=0.9pt,mark options={draw=white,line width=.2pt},keep={group}{Initialization}] table[col sep=comma,x=mflops,y=ne_delta_percent] {figures/frontier-prediction/observations-long.csv};
\addlegendentry{Initialization}
\addplot[only marks,black!65,mark=square*,mark size=1.3pt,mark options={draw=white,line width=.2pt},keep={group}{Adaptive}] table[col sep=comma,x=mflops,y=ne_delta_percent] {figures/frontier-prediction/observations-long.csv};
\addlegendentry{Adaptive}
\end{axis}
\end{tikzpicture}
\endgroup}
\caption{Frontier predictions made before the adaptive observations. Orange shows the fitted inverse-power curve; teal shows the GP median and pointwise central 90\% latent posterior band over 6,000 legal architectures, conditional on fitted hyperparameters, using 1,024 joint draws. The GP conditions on sixteen quality observations and 78 cost observations in total, including the sixteen initialization costs. Gray circles show the sixteen initialization observations and squares the four subsequent adaptive observations. Black steps show their empirical frontier. NE is relative to a shared baseline; lower is better.}
\label{fig:frontier-prediction}
\Description{NE versus inference FLOPs with an orange inverse-power curve and a teal GP median surrounded by its central ninety-percent pointwise posterior band. Gray circles and squares distinguish initialization and later adaptive observations; black steps show their measured frontier.}
\end{figure}

We compare two descriptions of the production parameterization's complexity--quality frontier, fitted before observing the four adaptive trials. An inverse-power curve, $\mathrm{NE}(F)=a+b/F^c$, fits the six observed-efficient initialization points. The frozen GP uses sixteen quality observations and 78 cost observations in total, including the sixteen initialization costs. Figure~\ref{fig:frontier-prediction} compares these predictions with the empirical frontier of the sixteen initialization and four subsequent adaptive observations. The predicted curves summarize opportunities within the search space; the empirical frontier records outcomes attained by the two-batch search.

\end{document}